\pdfoutput=1

\documentclass[11pt]{article}

\usepackage{acl}
\usepackage{times}
\usepackage{latexsym}
\usepackage[T1]{fontenc}

\usepackage[utf8]{inputenc}

\usepackage{microtype}

\usepackage{hyperref}
\usepackage{booktabs}
\usepackage{graphicx}
\graphicspath{{./imgs/}}
\usepackage{footmisc}
\usepackage{microtype}
\usepackage{arydshln}
\usepackage{caption}
\usepackage{subcaption}
\usepackage{array, makecell} %
\usepackage{footmisc}
\usepackage{alltt}
\usepackage{floatrow}
\usepackage{pifont}

\usepackage{tabularx}
\usepackage{adjustbox}
\usepackage{multirow}

\usepackage{multirow, colortbl, caption}
\definecolor{lightblue}{RGB}{232, 244, 248}
\definecolor{lightpink}{RGB}{254, 238, 237}

\definecolor{bluelink}{RGB}{0,113,188}
\definecolor{greenlink}{RGB}{0,188,113}
\definecolor{darkblue}{rgb}{0,0,0.55} %

\usepackage{tikz}

\usepackage{collcell}

\usepackage{etoolbox}

\newtoggle{inTableHeader}
\toggletrue{inTableHeader}
\newcommand*{\StartTableHeader}{\global\toggletrue{inTableHeader}}%
\let\OldTabular\tabular%
\let\OldEndTabular\endtabular%
\renewenvironment{tabular}{\StartTableHeader\OldTabular}{\OldEndTabular\StartTableHeader}%

\newcommand*{\MinNumber}{-1.0}%
\newcommand*{\MidNumber}{0.0} %
\newcommand*{\MaxNumber}{1.0}%

\newcommand{\ApplyGradient}[1]{%
  \iftoggle{inTableHeader}{#1}{
    \ifdim #1 pt > \MidNumber pt
        \pgfmathsetmacro{\PercentColor}{max(min(100.0*(#1 - \MidNumber)/(\MaxNumber-\MidNumber),100.0),0.00)} %
        \hspace{-0.33em}\colorbox{yellow!\PercentColor!blue}{#1}
    \else
        \pgfmathsetmacro{\PercentColor}{max(min(100.0*(\MidNumber - #1)/(\MidNumber-\MinNumber),100.0),0.00)} %
        \hspace{-0.33em}\colorbox{blue!\PercentColor!blue}{#1}
    \fi
  }}
\newcolumntype{R}{>{\collectcell\ApplyGradient}c<{\endcollectcell}}

\usepackage{amsmath}
\usepackage{amsfonts,bm}
\usepackage{xspace}

\newcommand{\Ni}{({\em i})~}
\newcommand{\Nii}{({\em ii})~}
\newcommand{\Niii}{({\em iii})~}
\newcommand{\Niv}{({\em iv})~}

\definecolor{mypink3}{cmyk}{0, 0.7808, 0.4429, 0.1412}

\makeatletter   
\newcommand{\sveryshortarrow}[1][3pt]{\mathrel{%
    \vcenter{\hbox{\rule[-.5\fontdimen8\scriptfont3]
               {\scriptratio\dimexpr#1\relax}{\fontdimen8\scriptfont3}}}%
   \mkern-4mu\hbox{\let\f@size\sf@size\usefont{U}{lasy}{m}{n}\symbol{41}}}}
\makeatother

\def\eqref#1{equation~\ref{#1}}

\def\1{\bm{1}}

\def\m1{{\bm{1}}}

\DeclareMathAlphabet{\mathsfit}{\encodingdefault}{\sfdefault}{m}{sl}
\SetMathAlphabet{\mathsfit}{bold}{\encodingdefault}{\sfdefault}{bx}{n}

\usepackage[nameinlink]{cleveref}
\crefformat{section}{\S#2#1#3} 
\crefname{algorithm}{Alg.}{Algs.}
\crefname{table}{Table}{Tables}
\crefformat{subsection}{\S#2#1#3}
\Crefname{equation}{Eq.}{Eqs.}
\Crefname{figure}{Figure}{Figures}

\usepackage[colorinlistoftodos,prependcaption,textsize=tiny]{todonotes}

\definecolor{darkgreen}{rgb}{0,0.5,0}  

\usepackage{soul}

\usepackage{float}

\usepackage{multirow}
\usepackage{hhline}

\definecolor{chartqapro1}{RGB}{30,160,220} 
\definecolor{chartqapro2}{RGB}{50,200,100} 

\title{
\textbf{Text2Vis: A Challenging and Diverse Benchmark for Generating Multimodal Visualizations from Text}
}

 \author{Mizanur Rahman\textsuperscript{\textdaggerdbl,}\thanks{\hspace{0.115cm} Contact Emails: \{mizanurr,enamulh\}@yorku.ca},    \textbf{Md Tahmid Rahman Laskar}\textsuperscript{\textdaggerdbl,\textsection}\textbf{,}  
 \textbf{Shafiq Joty\textsuperscript{\textparagraph}}\textbf{, }
 \textbf{Enamul Hoque\textsuperscript{\textdaggerdbl,}}\footnotemark[1]\textbf{, } \\
            {\textsuperscript{\textdaggerdbl}York University, \textsuperscript{\textsection}Dialpad, \textsuperscript{\textparagraph}Salesforce AI Research}
          \\ }
\begin{document}
\maketitle

\begin{abstract} 
Automated data visualization plays a crucial role in simplifying data interpretation, enhancing decision-making, and improving efficiency. While large language models (LLMs) have shown promise in generating 
visualizations from natural language, the absence of comprehensive benchmarks limits the rigorous evaluation of their capabilities. We introduce Text2Vis, a benchmark designed to assess text-to-visualization models, covering 20+ chart types and diverse data science queries, including trend analysis, correlation, outlier detection, and predictive analytics. It comprises 1,985 samples, each with a data table, natural language query, short answer, visualization code, and annotated charts. The queries involve complex reasoning, conversational turns, and dynamic data retrieval. We benchmark 11 open-source and closed-source models, revealing significant performance gaps, highlighting key challenges, and offering insights for future advancements. To close this gap, we propose the first cross-modal actor-critic agentic  framework that jointly refines the textual answer and visualization code, increasing GPT-4o’s pass rate from 26\% to 42\% over the direct approach and improving chart quality. We also introduce an automated LLM-based evaluation framework that enables scalable assessment across thousands of samples without human annotation, measuring answer correctness, code execution success, visualization readability, and chart accuracy. 
We release Text2Vis at \url{https://github.com/vis-nlp/Text2Vis}.

\end{abstract}

\section{Introduction}

\begin{table*}[t!]
\centering
\renewcommand{\arraystretch}{1.2} 
\small
\setlength{\tabcolsep}{3pt} 

\caption{
Comparison of \textbf{Text2Vis} with existing text-to-visualization benchmarks.}
\label{tab:datasets_comparison}

\resizebox{0.9\textwidth}{!}{ 
\begin{tabular}{>{\raggedright\arraybackslash}l c c c c c c c c c c c}
\toprule
\textbf{Dataset} & \makecell{\textbf{Data} \\ \textbf{Type}} & \makecell{\textbf{Query} \\ \textbf{Type}} & \makecell{\textbf{Web Data} \\ \textbf{Retrieval}} & \makecell{\textbf{Conversa}-\\ \textbf{tional}} & \makecell{\textbf{Unans}- \\ \textbf{werable}} & \makecell{\textbf{Multistep} \\ \textbf{reasoning}} & \makecell{\textbf{Text} \\ \textbf{Annotations}} & 
\makecell{\textbf{Text} \\ \textbf{Explainability}} 
 & \makecell{\textbf{Chart} \\ \textbf{Variety}} & \makecell{\textbf{NLP to} \\ \textbf{Python}} & \makecell{\textbf{Chart} \\ \textbf{Specification}} \\ 
\midrule
\rowcolor[HTML]{F2F2F2} WikiSQL~\cite{zhong2017seq2sql} & Real & NL2SQL & \ding{55} & \ding{55} & \ding{55} & \ding{55} & \ding{55} & \ding{55} & SQL Only & No & N/A \\ 
\rowcolor[HTML]{F2F2F2} nvBench~\cite{luo2021nvbench} & Synthetic & NL2Vis & \ding{55} & \ding{55} & \ding{55} & \ding{55} & \ding{55} & \ding{55} & 7 Types & No & Direct \\ 
\rowcolor[HTML]{F2F2F2} NLV-Utterance~\cite{srinivasan2021collecting} & Real & Simple Agg. & \ding{55} & \ding{55} & \ding{55} & \ding{55} & \ding{55} & \ding{55} & 10 Types & No & Direct \\ 
\rowcolor[HTML]{F2F2F2} ADVISor~\cite{liu2021advisor} & Real & Aggregation & \ding{55} & \ding{55} & \ding{55} & \ding{55} & \ding{51} & \ding{55} & 3 Types & No & Direct \\ 
\rowcolor[HTML]{F2F2F2} VisEval~\cite{chen2024viseval} & Real + Synth. & Mid-Complex & \ding{55} & \ding{55} & \ding{55} & \ding{51} & \ding{55} & \ding{55} & 7 Types & Yes & Direct \\ 
\midrule
\rowcolor[HTML]{E5F1FB} \textbf{Text2Vis (Ours)} & Real + Synth. & Complex Hard & \ding{51} & \ding{51} & \ding{51} & \ding{51} & \ding{51} & \ding{51} & \textgreater 20 Types & Yes & Open \\ 
\bottomrule
\end{tabular}
} 
\vspace{-3mm}
\end{table*}

 \begin{figure}[t!]
    \includegraphics[width=\textwidth]{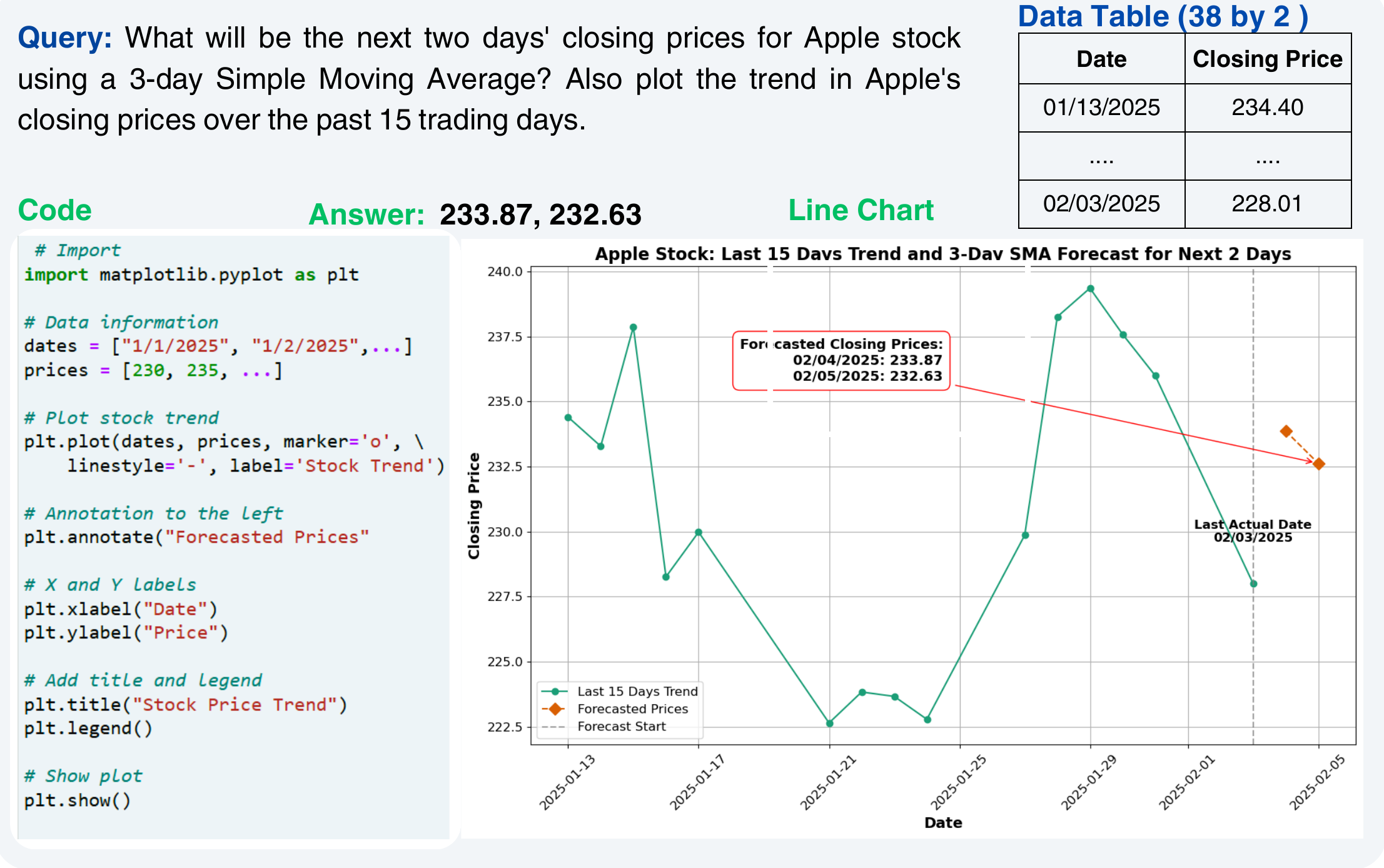}
    \caption{\textbf{Example from the Text2Vis benchmark.} 
    \textcolor{darkblue}{\textbf{Input:}} A data table containing historical stock prices and a query.  
    \textcolor{darkgreen}{\textbf{Output:}} Python code for visualization, the predicted answer, and an annotated textual explanation. The chart is generated from the code.}
    \label{fig:performance_gap}
    \vspace{-2mm}
\end{figure}

Data visualization transforms raw data into meaningful visual representations, allowing users to gain insights and make data-driven decisions 
\cite{aparicio2015data, hoque2022chartquestionansweringstate}. It is an integral part of the data science workflow, frequently used for exploratory data analysis, outlier detection, pattern recognition, and feature identification. However, creating accurate and intuitive visualizations is challenging due to the need to correctly interpret natural language queries, locate and, if necessary, transform the relevant data, identify the appropriate chart type, and generate the correct visualization code~\cite{shen2022towards}. This problem 
integrates multiple modalities—visual representation, natural language understanding, logical reasoning, and code generation—making it
more challenging 
than traditional NLP tasks. Moreover, this process typically requires expertise in data science, programming languages, and visualization libraries (e.g., Matplotlib, Vega-Lite~\cite{satyanarayan2016vega}), creating a significant barrier for non-technical users \cite{ali2016big, waskom2021seaborn, bisong2019matplotlib}. While natural language interfaces (NLIs) like Tableau’s Ask Data~\cite{askdata} could be helpful for non-technical users as these NLI platforms can generate basic charts from queries, they lack flexibility, automation, customization, and advanced analytical capabilities.

LLMs have demonstrated strong performance in code generation and data analysis~\cite{nejjar2025llms}, making them promising for automated visualization tasks \cite{liu2021advisor, hoque2024natural, maddigan2023chat2vis}. In addition to LLMs understanding natural language queries, they can also identify the relevant data attributes and the appropriate chart types for visualization.  
By generating visualization code, 
LLMs can lower the barrier for non-experts to explore data without extensive technical expertise. 

However, as LLMs advance in coding and reasoning—often rivaling human performance on benchmarks—there is a growing need for more rigorous evaluations with complex, real-world tasks. Existing text-to-visualization benchmarks often fail to capture this complexity,  limiting themselves primarily to natural language to SQL translation or basic visualizations that map explicitly mentioned data columns to chart elements~\cite{zhong2017seq2sql, luo2021nvbench, srinivasan2021collecting, liu2021advisor}.  In addition, most rely on rule-based approaches or declarative specifications like Vega-Lite, which restrict flexibility and hinder support for advanced customization or logic, unlike executable code generation. While the recent
VisEval benchmark~\cite{chen2024viseval} evaluates LLM-driven visualization generation, it lacks diverse chart types, real-world data science tasks, and multi-step reasoning.  Moreover, most queries in this benchmark have explicit chart-type mentions (e.g., “draw a line chart...”)  rather than accommodating open-ended queries.

Another major limitation of existing benchmarks is their omission of concise textual answers alongside generated visualizations, despite the fact that users often create charts to address specific data-driven questions. For example, as shown in Figure~\ref{fig:performance_gap}, a query asking for a 3-day moving average to predict a stock’s closing price benefits not only from the generated chart, but also from an answer that clarifies the logic used. Such multimodal outputs—visual and textual—are essential for robust, interpretable systems.

\begin{figure*}[t]
    \vspace{-3mm}
    \centering
    \includegraphics[width=\textwidth]{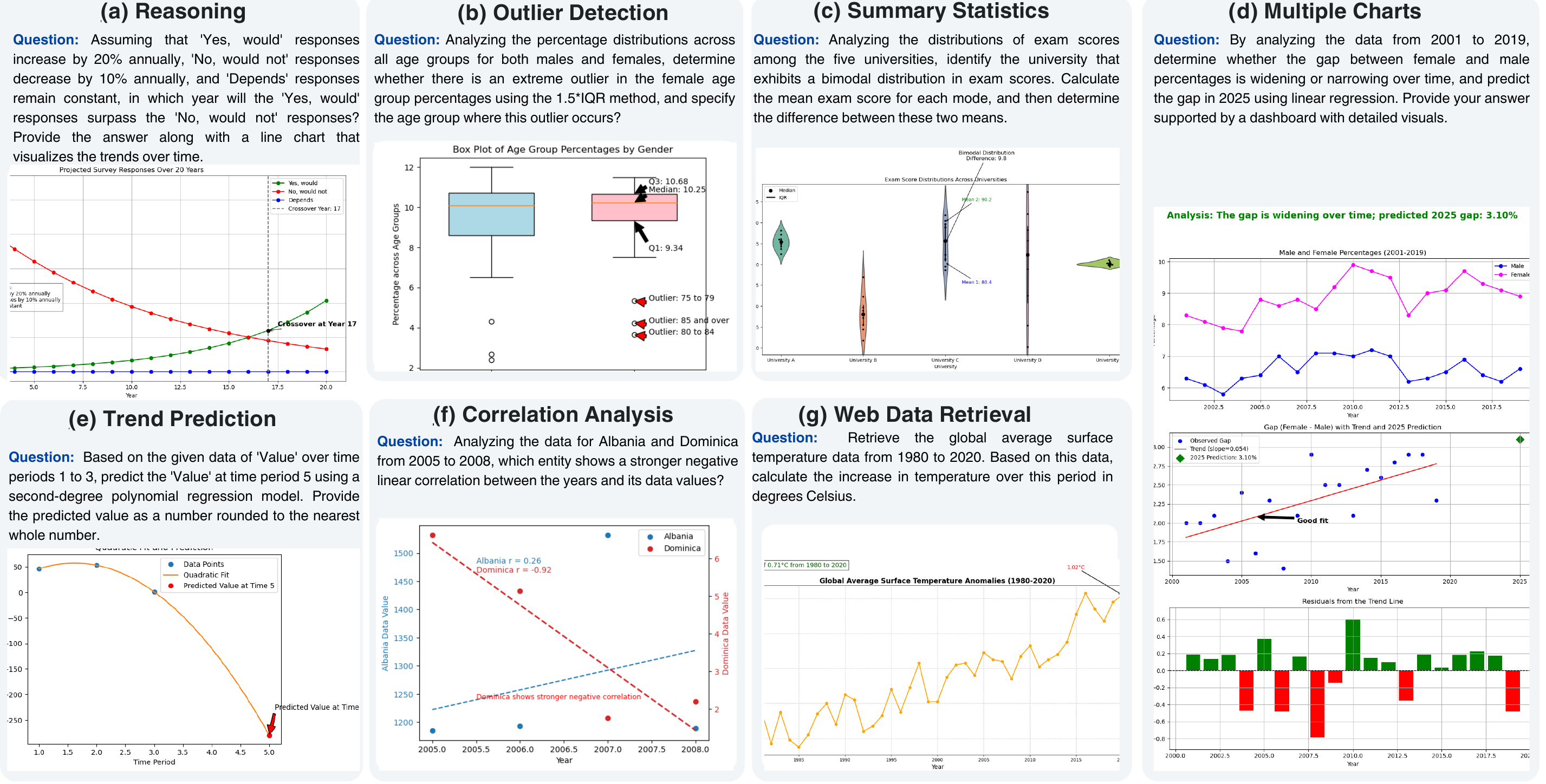}
    \caption{Examples of different question types used in data analysis, including trend prediction, reasoning, outlier detection, correlation analysis, summary statistics, and retrieval-augmented tasks.
    }
    \label{fig:qa_types}
    \vspace{-3mm}
    \vskip -.5ex
\end{figure*}

To address these gaps, we introduce Text2Vis, a comprehensive benchmark for evaluating LLMs on real-world text-to-visualization tasks. It features 1,985 diverse samples, each comprising a data table, a natural language query, a short answer, visualization code, and annotated charts. Unlike previous benchmarks, Text2Vis covers over 20 chart types and supports a wide range of realistic analytical scenarios, including retrieval-augmented queries, multi-turn conversations, multi-chart outputs, and unanswerable questions (Table \ref{tab:datasets_comparison}). It also covers complex reasoning tasks such as statistical analysis, trend forecasting, outlier detection (Figure \ref{fig:qa_types}).

To advance LLM capabilities, we propose a cross-modal actor-critic inference framework that jointly refines textual answers and visualizations using multimodal feedback. This improves answer accuracy and chart quality, outperforming direct inference. We also introduce a fully automated LLM-based evaluation framework to assess answer and chart correctness, chart readability, and visual accuracy—validated on 1,985 samples across 11 models, closely aligning with human judgments and eliminating the need for manual annotation.


In summary, our contributions include: \Ni  \textbf{Text2Vis}, a comprehensive benchmark featuring 1,985 queries that reflect diverse, real-world data science challenges, including complex analytical reasoning and multi-step tasks; \Nii a cross-modal \textbf{actor-critic agentic inference framework} that simultaneously refines both answer and visualization code, which significantly enhances the accuracy, readability, and reliability of generated outputs; \Niii a scalable \textbf{LLM-based  evaluation framework} that systematically assesses answer correctness, visualization readability, and chart accuracy—enabling consistent, large-scale benchmarking without human annotation; 
and \Niv \textbf{extensive evaluations} with 11 open- and closed-source models, revealing significant performance gaps and common failure patterns, providing valuable directions for future research.

\vspace{-2mm}

\section{Related Work}
\textbf{Text-to-Visualization Benchmarks} Existing text-to-visualization benchmarks often oversimplify real-world analytical tasks by framing them as NL-to-SQL translation—assuming visualizations can be derived from SQL outputs—or by reducing the task to visualization specification mapping \cite{zhong2017seq2sql, luo2021nvbench, srinivasan2021collecting, liu2021advisor, chen2024viseval}. For example, WikiSQL \cite{zhong2017seq2sql} and nvBench \cite{luo2021nvbench} focus primarily on NL2SQL tasks, while NLV-Utterance \cite{srinivasan2021collecting} and ADVISor \cite{liu2021advisor} map textual queries to visualization specifications but do not support complex analytical queries or multi-step reasoning.  Most rely on Vega-Lite rather than Python code, limiting practical use in customizable workflows. 
VisEval \cite{chen2024viseval}, adapted from nvBench, is constrained by its small set of tables (146), limited chart variety, and explicit chart-type mentions in queries, reducing its real-world relevance.

As summarized in Table~\ref{tab:datasets_comparison}, current benchmarks have three core limitations: (1) limited coverage of questions and chart types, (2) lack of multi-step analytical reasoning, and (3) weak alignment with real-world workflows such as web data retrieval, conversational input, and unanswerable queries. These gaps motivate the need for a more comprehensive benchmark, which we present in this work.
\noindent \textbf{LLMs for Automated Visualization}
Visualization generation has progressed from rule-based templates to deep learning and LLMs. Early methods relied on predefined templates but struggled with ambiguity and scalability \cite{narechania2020nl4dv}, leading to hybrid approaches like RGVisNet \cite{song2022rgvisnet} and ADVISor \cite{liu2021advisor}, which improved data extraction and visualization generation. Recent LLMs have significantly advanced code generation capabilities \cite{hoque2024natural,maddigan2023chat2vis}. Chat2VIS \cite{maddigan2023chat2vis} applied prompt engineering for visualizations, while ChartLlama \cite{han2023chartllama}  enhanced chart understanding through instruction tuning.  Yet, challenges in grounding, correctness, and execution persist \cite{chen2024viseval}. We address these gaps with a cross-modal agentic inference framework that jointly refines answers and code using multimodal feedback.

\noindent \textbf{Visualization Evaluation} Early works like ADVISor~\cite{liu2021advisor} and NLV-Utterance~\cite{srinivasan2021collecting} focused on verifying syntactic correctness and manually inspecting visualizations but lacked scalability for complex queries and large datasets. \citet{chen2023beyond} used LLMs for interpreting data and designing visualizations, but relied on manual evaluation. \citet{podo2024vi} proposed a structured framework assessing code correctness, chart legality, and semantic alignment. VisEval~\cite{chen2024viseval} partially leveraged LLMs, using GPT-4V only for readability scoring, while relying on rule-based checks for code validity and chart legality. 
 Although LLMs show high agreement with human judgments \cite{gu2024survey}, their broader use in evaluating visualizations remains limited. We propose a fully automated LLM-based framework for assessing answer and chart correctness, visual quality, and readability, enabling scalable benchmarking with minimal human effort.
\vspace{-2mm}
\section{\textsc{Text2Vis}}



\vspace{-2mm}
We curated and synthesized a diverse dataset of data tables, queries, charts, and metadata. The dataset creation involved three key steps: (1) data table collection, (2) query generation and annotation, and (3) dataset analysis.
\vspace{-2mm}
\subsection{Data Table Construction}
We started with the existing ChartQA corpus~\cite{masry2022chartqa}, which originally scraped 22K data tables from four diverse sources: 
\Ni Statista~\cite{statista}, \Nii Pew Research ~\cite{pewresearch}, \Niii Our World In Data or OWID~\cite{pewresearch}, and \Niv Organisation for Economic Co-operation and Development or OECD ~\cite{owid}.
From this collection, we manually curated 2001 high-quality tables based on complexity, diversity, and analytical richness. 

To broaden the dataset variety and increase complexity further, we generated 173 synthetic tables using OpenAI o1-preview and Gemini Flash 1.5 Pro (Table~\ref{tab:prompt_template_synthetic}), incorporating missing values, multi-variable dependencies, and non-linear patterns.  
\subsection{{Query Generation and Annotations}}  
\paragraph{Query Generation and Expansion}  
Three co-authors of this paper, who are also experts in data science, manually crafted 600 high-quality queries reflecting real-world challenges such as trend analysis, statistical computations, correlation analysis, outlier detection, comparisons, deviation analysis, predictive modeling, time-series analysis, forecasting, and geospatial analysis. These queries emphasize complex reasoning, making them more challenging than those in existing benchmarks.  To expand this initial set, we leveraged multiple LLMs, including OpenAI o1-preview, Gemini Flash 1.5 Pro, and Claude 3.5 Haiku. Using few-shot prompting, we generated 1,624 additional queries 
(see Table~\ref{tab:prompt_templat_conv}), broadening the coverage of analytical tasks and reasoning-based challenges. After manual verification, 239 table-query pairs were removed due to issues with table quality or overly simple queries. The final dataset comprises 1,985 samples: 1,935 based on curated tables, and 50 designed for web-based data retrieval, providing a robust benchmark for evaluating text-to-visualization models across 
complex analytical tasks in real-world scenarios.  
\paragraph{Visualization Code and Answer Generation}  
For each query, we generated visualization code using OpenAI o1-preview based on two libraries, Matplotlib \cite{bisong2019matplotlib} and Seaborn~\cite{waskom2021seaborn}, as these are among the most versatile and widely used data visualization libraries in Python. In addition, we generated short answers, visualization summaries, and metadata, including chart type and axis labels. All outputs were manually reviewed, corrected, and refined to ensure accuracy, clarity, and relevance.
\begin{table*}[th]
\vspace{-3mm}
\centering
\renewcommand{\arraystretch}{1.0} 
\setlength{\tabcolsep}{5pt}     

\caption{\small Distribution of question categories, chart types, question types based on complexity, and tasks type in Text2Vis.}
\label{tab:data_diversity}

\resizebox{\textwidth}{!}{%
\begin{tabular}{c c c c c c c c c c c c c}
\toprule
\multicolumn{5}{c}{\textbf{Question Category (\%})} & \multicolumn{4}{c}{\textbf{Question Complexity}} & \multicolumn{3}{c}{\textbf{Task Type}} \\
\cmidrule(lr){1-5} \cmidrule(lr){6-9} \cmidrule(lr){10-13}
\makecell{ Closed/\\ Open-Ended} & \makecell{Single query/\\ Conversational} & \makecell{Data Given/\\Web-data Retrieval} & \makecell{Single/\\Multi-Chart} & \makecell{Answerable/\\ Unanswerable} & Easy & Medium & Hard & Extra Hard & Analytical & Exploratory & Predictive & Prescriptive\\
\midrule
\rowcolor[HTML]{E5F1FB}
90/10 & 80/20 & 97/3 & 90/10 & 89/11 & 343 & 245 & 1173 & 224 & 1098 & 686 & 191 & 10 \\
\bottomrule
\end{tabular}%
}
\vspace{-3.5mm}
\end{table*}
\vspace{-1mm}

\subsection{Dataset Analysis}

\textbf{Data Table Statistics} 
Our dataset consists of 1,985 data samples covering over 60 countries and diverse demographic and sectoral domains, including finance, healthcare, politics, energy, technology, demographics, and environment. It exhibits structural diversity, with tables containing an average of 10 rows (max: 1,000) and 3.2 columns (max: 15), ensuring a mix of compact and extensive datasets. In addition to clean data tables, the dataset also contains noisy tables (191 tables) featuring missing values or inconsistencies, as well as hybrid cases, enabling robust evaluation of models handling real-world inconsistencies.


\noindent\textbf{Query Diversity}  
To characterize query types, we used GPT-4o to automatically categorize each natural language query across three dimensions: \Ni Question type,  \Nii Question complexity and \Niii Task type. See Appendix (Table~\ref{tab:prompt_template_complexity}) for an example prompt used in complexity classification.

Text2Vis supports a diverse set of question types that evaluate various aspects of analytical reasoning and visualization generation (Table~\ref{tab:data_diversity}). While most questions take a given data table and query as input, expecting a specific answer as output (closed-ended), others are open-ended, allowing for multiple possible visualizations \Cref{app:query-types}. 20\% questions involve multi-turn conversations, simulating natural dialogue in analytical workflows. Similarly, while many questions provide data tables, a small number of queries (3\%) require models to retrieve external data before generating visualizations. Additionally, certain questions expect models to produce multiple visualizations to explore complex datasets (10\%), reflecting real-world scenarios in dashboards and infographics where a single visualization is insufficient. Finally, unanswerable queries (11\%) appear across all categories, adding complexity by requiring models to recognize when a valid response cannot be generated. The overall query set is highly challenging, with most questions categorized as hard (1,173) or extra hard (224). Examples of different query types are illustrated in \Cref{fig:qa_types} and \Cref{app:query-types}.

The dataset spans a broad range of data science tasks, including analytical (1098 queries), exploratory (686), predictive (191), and prescriptive (10), capturing real-world multi-step and interactive data exploration scenarios (Table~\ref{tab:data_diversity}). It also demonstrates significant linguistic richness, with an average question length of 217.87 characters and 34.15 tokens, covering a vocabulary of 6,776 unique tokens. This ensures syntactic complexity and variability. The combination of diverse data sources, multi-faceted queries, and linguistic depth makes Text2Vis a challenging and realistic benchmark for text-to-visualization models.

\noindent\textbf{Code Diversity and Complexity} Matplotlib and Seaborn are two of the most widely used Python libraries for visualization, and we provide code in both to ensure broad compatibility and adaptability. To measure the diversity of axis labels, we used cosine distance between the TF-IDF vectors of the axis labels (for both X and Y). The average distance was 0.97, indicating 
that our dataset includes a wide range of unique labels, covering different contexts and visualization types. 
In terms of code complexity, our scripts average 33.74 lines of code, 1,146 characters, and 123.72 tokens. Additionally, with an average of 5.34 comments per script, we prioritize clarity and maintainability, aligning with real-world visualization coding practices.



\noindent\textbf{Visual Diversity}  
Our dataset includes over 20 types of visualizations, covering not only common charts like bar and line charts but also more complex and less frequent types such as treemaps, boxplots, waterfall charts, and dashboard-style multi-chart visualizations (\Cref{fig:chart_types}). This diverse collection enhances model robustness by exposing it to a wide range of chart types and visual styles (e.g., color, layout).
To quantify color diversity, we converted images to LAB color space and computed pairwise Euclidean distances between dominant colors, yielding a Mean CIEDE2000 \cite{sharma2005ciede2000} Color Distance of 13.9, indicating strong variation in color schemes. For multi-modal feature analysis, we used OCR to extract chart text, derived visual features using CLIP (Contrastive Language-Image Pretraining) \cite{radford2021learning} - whose effectiveness for representing chart images was validated via a small controlled experiment, and computed text embeddings via a Sentence Transformer \cite{reimers2019sentence}.  This produced a Mean Cosine Distance of 0.69, highlighting strong diversity in chart structures and annotations.

\vspace{-2mm}
\section{Methodology}
\vspace{-1mm}


\subsection{Problem Formulation}
We define the Text2Vis task as a text-to-visualization generation problem that evaluates how well a model can translate natural language queries into a visualization annotated with concise textual answers. The dataset consists of \( N \) examples, denoted as \( D = \{t_i, q_i, a_i, v_i\}_{i=1}^{N} \), where each example includes a data table \( t_i \), a natural language query \( q_i \), the corresponding short answer \( a_i \), and the visualization code \( v_i \). The model is tasked with generating both \( a_i \) and \( v_i \) based on \( t_i \) and \( q_i \), with \( v_i \) producing an executable visualization code. 

\vspace{-1mm}
\subsection{Models} \label{subsec:data2text}

We evaluated both state-of-the-art closed-source models and open-source models to benchmark text-to-visualization generation capabilities. For closed-source models, we tested GPT-4o \cite{openai2024gpt4technicalreport} and Gemini 1.5-Flash \cite{geminiteam2024gemini15unlockingmultimodal}, which are widely used for natural language understanding and code generation.  For open-source models, we prioritized deployment feasibility in the real-world and mostly selected models with less than 10B parameters. More specifically, we evaluated Qwen2.5-7B-Instruct, Qwen2.5-7B-Coder \cite{yang2024qwen2}, Mistral-7B \cite{jiang2023mistral}, LLaMA 3.1-8B \cite{grattafiori2024llama}, DeepSeek-Coder-V2-Lite \cite{DBLP:journals/corr/abs-2406-11931} and DeepSeek-R1-Distill-LLaMA-8B \cite{guo2025deepseek}, as well as
CodeLlama-7B-Instruct \cite{roziere2023code}. For CodeLlama, we also use its 13B and 34B versions. 

 \vspace{-1mm}
\subsection{Text2Vis Inference Approaches} \label{subsec:evalmethod}
\begin{figure}[t!]
    \centering
    \includegraphics[width=\textwidth]{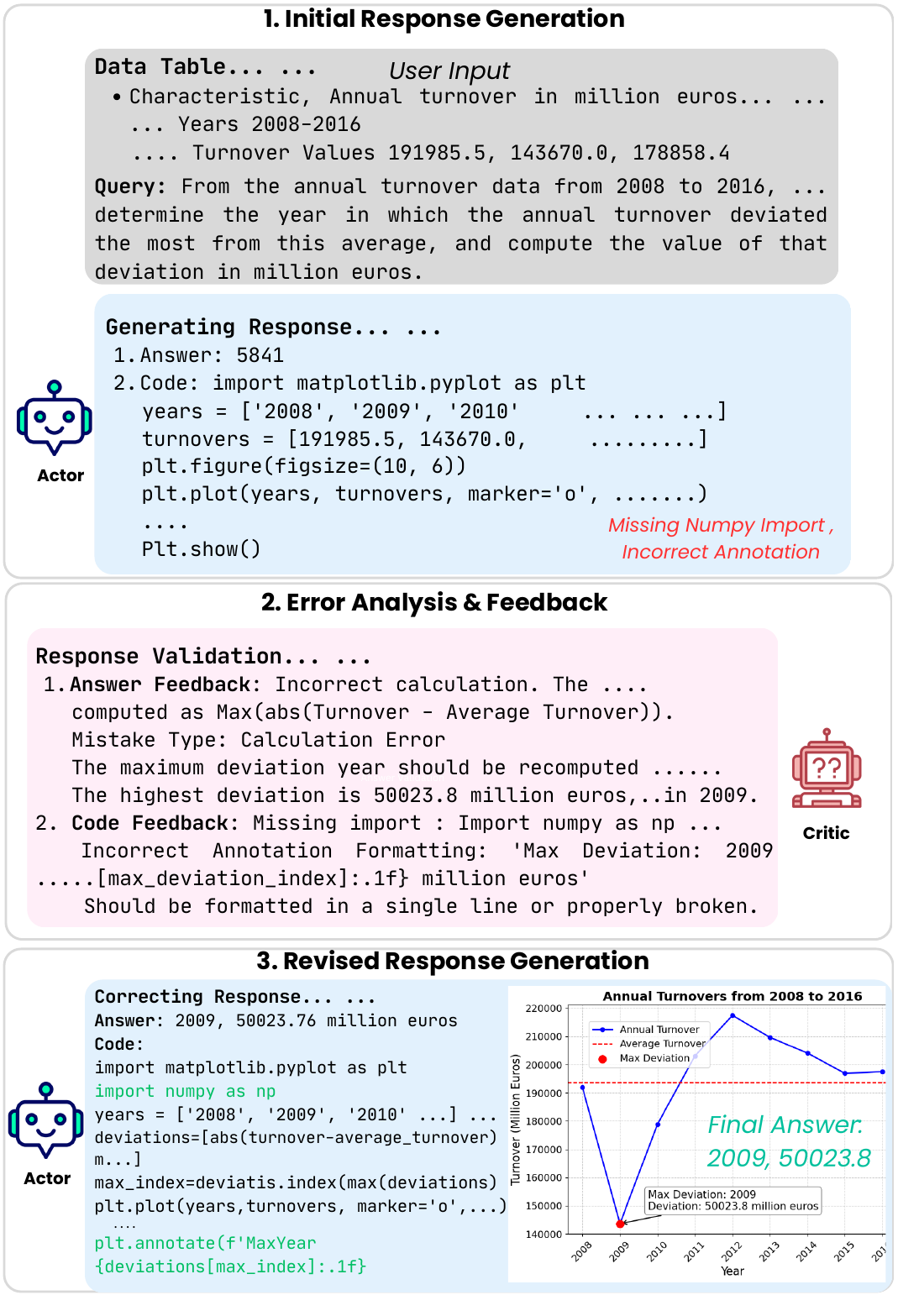}
    \caption{Our Agentic Inference Framework where the Actor (e.g., Gemini) generates an initial response, while the Critic (e.g., GPT-4o for validation, Matplotlib for visualization execution) assesses and provides feedback. 
    }
    \vspace{-2mm}
    \label{fig:agentic}
\end{figure}

We use two approaches to assess the performance of text-to-visualization models: a direct inference and an agentic inference framework (Fig \ref{fig:agentic}). 


\noindent \textbf{(i) Direct Inference:}
In this method, the model is given a prompt containing a natural language query, a data table and instructions to generate a JSON response containing both a short answer and visualization code. We evaluate three prompting strategies: (a) zero-shot prompting, (b) 3-shot prompting, and (c) retrieval-augmented 3-shot prompting with dynamically selected examples  (\Cref{app:direct-inference-setups}).  

\noindent \textbf{(ii) Agentic Inference:}
We propose a cross-modal actor-critic inference framework, inspired by reflective LLM workflows \cite{islam2024datanarrative,DBLP:conf/nips/ShinnCGNY23, madaan2023self}.  To our knowledge, this is the first agentic setup for text-to-visualization that incorporates multimodal feedback on answer correctness, code quality, and the resulting visualization  (Figure \ref{fig:agentic}). The framework is also model-agnostic: it takes the initial $\langle \text{answer}, \text{code} \rangle$ output from any baseline inference model $f_{\theta}$ and routes it into the actor--critic refinement loop.
The key steps are:

\textbf{(1) Initial Response Generation (Actor Step):}
The actor model generates an initial response containing the answer and visualization code based on the given query and data table. 

\textbf{(2) Critic Evaluation \& Feedback Generation:} A separate critic model analyzes the initial response and provides structured feedback across three modalities: answer feedback (numerical correctness), code feedback (syntax/semantic checks), and visual feedback (clarity and correctness of the generated chart). The visual feedback is 
for the chart 
produced by executing the generated code.


\textbf{(3) Refinement \& Final Response Generation:}
The actor takes both the initial response and the critic's feedback into account to produce a refined final response. This iterative refinement process ensures that the final output is more aligned 
with the intent of the query. To ensure inference efficiency, only one round of iteration is performed. 


\noindent\textbf{Feedback Strategies.} We explore three critic configurations: (1) self-critique (same model) \cite{DBLP:journals/corr/abs-2206-05802}, (2) cross-model critique (external model), and (3) execution-based feedback (e.g., Matplotlib error traces). Each variant affects the reliability of downstream refinement.


\vspace{-1mm}
\subsection{Evaluation Criteria} \label{subsec:evalmetric}
To comprehensively assess text-to-visualization models, we define four key evaluation criteria. 

\textbf{Answer Match}: Assesses how well the generated text aligns with the ground truth.


\textbf{Code Execution}: Measures whether the generated visualization code executes successfully, ensuring syntactic correctness and output generation.

\textbf{Readability and Visualization Quality}: Assesses the visual clarity and design quality of the chart, including layout, axis scaling, labels, titles, and color usage~\cite{chen2024viseval}. 

\textbf{Chart Correctness}: Measures whether the generated chart accurately represents the intent of the query and the underlying data. 

 Scoring is binary for Answer Match and Code Execution (1 = success, 0 = failure), while Readability and Chart Correctness are rated from 1 to 5 (Table \ref{tab:result_evaluation}). A sample is considered a pass if the code executes successfully, the answer matches the ground truth, and both the readability score and the chart correctness score are at least 3.5, implying that minor readability or correctness issues may exist, but the output remains interpretable.
\vspace{-5mm}
\paragraph{Automatic Judge:} We use GPT-4o as the evaluation judge for all criteria, including Answer Match, Readability, Visualization Quality, and Chart Correctness, based on its strong alignment with human judgments~\cite{zheng2023judging,gu2024survey}. The model is invoked with a fixed system prompt and deterministic parameters (see~\ref{app:exp-setup}). The evaluation rubric and prompt are detailed in Table~\ref{tab:result_evaluation}. To demonstrate scalability, we evaluated 1,985 samples in 5 minutes for a total cost of ~\$2.0 (see  \ref{app:cost}).


\vspace{-2mm}

\section{Experiment Results}
\label{sec:benchmark_results}
\vspace{-1mm}
\subsection{Automatic  Evaluation}
\paragraph{ Results for Direct Inference:} 
Table~\ref{tab:text2vis_dual_finalrate} presents the automated evaluation results for all models assessed using the direct inference approach. The models were evaluated based on predefined criteria. GPT-4o achieved the highest performance, with 87\% code execution success, 42\% correct answer match, average visual clarity rating of 3.45, chart correctness of 3.15, and a final pass rate of 26\%. 
Among open-source models, Qwen2.5-7B performed the best, followed closely by DeepSeek-Coder-V2-Lite, both achieving a final pass rate of 13\% and 10\% respectively.

\definecolor{human_baseline}{RGB}{255, 245, 170}
\definecolor{open_models_below_4B}{RGB}{185, 235, 255}
\definecolor{open_models_7B_12B}{RGB}{255, 219, 187}
\definecolor{closed_models}{RGB}{240, 240, 240}
\definecolor{chart_specific_models}{RGB}{217, 240, 211}

\begin{table}[t!]
\vspace{-1mm}
\centering
\small
\setlength{\tabcolsep}{1.7pt}   
\renewcommand{\arraystretch}{1.25} 

\caption{Automatic evaluation on Text2Vis using direct inference across models. Higher scores indicate better performance. Visual Clarity, Readability, and Correctness are rated out of 5 by GPT-4o.  The last column gives Final Pass Rate from Gemini 1.5 Pro for comparison.}

\label{tab:text2vis_dual_finalrate}

\resizebox{1\columnwidth}{!}{%
\begin{tabular}{>{\raggedright\arraybackslash}p{4.2cm}|c|c|c|c|cc}
\hline
\textbf{Model} &
\makecell{\textbf{Code Exec.}\\\textbf{Success (\%)}} &
\makecell{\textbf{Answer}\\\textbf{Match (\%)}} &
\makecell{\textbf{Visual Clarity}\\\textbf{Readability}} &
\makecell{\textbf{Chart}\\\textbf{Correctness}} &
\multicolumn{2}{c}{\makecell{\textbf{Final}\\\textbf{Pass Rate (\%)}}} \\
\cline{6-7}
& & & & & \textbf{GPT} & \textbf{Gemini} \\
\hline
\rowcolor[HTML]{E5F1FB}\textbf{GPT-4o}                  & \textbf{87} & \textbf{42} & \textbf{3.45} & \textbf{3.15} & \textbf{26} & \textbf{24} \\
\rowcolor[HTML]{E5F1FB}Gemini-1.5-Flash                 & 83 & 34 & 3.30 & 2.90 & 17 & 15 \\
\rowcolor[HTML]{E5F1FB}CodeLlama-7B                     & 60 & 10 & 2.15 & 1.69 &  1 & 1 \\
\rowcolor[HTML]{E5F1FB}CodeLlama-13B                    & 52 & 15 & 1.75 & 1.38 &  4 & 3 \\
\rowcolor[HTML]{E5F1FB}CodeLlama-34B                    & 39 & 22 & 0.91 & 0.80 &  4 & 3 \\
\rowcolor[HTML]{E5F1FB}Llama-3.1-8B                     & 72 & 24 & 1.68 & 1.59 &  7 & 6 \\
\rowcolor[HTML]{E5F1FB}Mistral-7B                       & 39 & 24 & 1.40 & 1.31 &  6 & 4 \\
\rowcolor[HTML]{E5F1FB}Qwen-2.5-7B                      & 80 & 29 & 2.82 & 2.73 & 13 & 12 \\
\rowcolor[HTML]{E5F1FB}Qwen-2.5-Coder-7B                & 31 & 24 & 1.25 & 1.26 &  4 & 3 \\
\rowcolor[HTML]{E5F1FB}DeepSeek-Coder-V2-Lite           & 75 & 22 & 2.93 & 2.63 & 10 & 8 \\
\rowcolor[HTML]{E5F1FB}DeepSeek-R1-Distill-Llama-8B     & 35 & 33 & 1.24 & 1.12 &  7 & 8 \\
\hline
\end{tabular}}
\end{table}

Despite its larger size, CodeLlama-34B performed poorly, reinforcing that increased model size does not necessarily improve structured data comprehension. Over 50\% of its failures were due to incorrect extraction of relevant data elements, leading to execution errors. 
\vspace{-2mm}
\paragraph{Results for Agentic Inference:}
{Table~\ref{tab:gpt4o_agentic_performance} shows that our agentic framework consistently improves GPT-4o’s performance across all metrics, outperforming zero-shot, few-shot, and RAG baselines. With one round of Answer + Code feedback, the answer match improved from 42\% to 53\%, readability from 3.45 to 3.99, chart correctness from 3.15 to 4.02, and the final pass rate from 26\% to 42\% (+62\%). Adding visual feedback further improved readability (4.23) and chart correctness (4.24), though answer match remained stable—indicating its impact is primarily visual. Among all settings, Answer + Code feedback yielded the strongest final pass rate.  In contrast, code-only and execution-only feedback achieved high execution rates but low final pass rates due to poor answer accuracy. Also, all improvements in final pass rate over the zero-shot baseline are statistically significant ($p < 0.01$), based on McNemar’s test on all 1,985 paired samples.
These findings highlight the benefit of multimodal feedback in agentic refinement.

\begin{table}[t!]
\vspace{-1.5mm}
\centering
\setlength{\tabcolsep}{1.7pt} 
\renewcommand{\arraystretch}{1.25}
\caption{
Model performance for \textbf{(A) Baseline}, \textbf{(B) Agentic Inference}, and \textbf{(C) LLM Feedback Ablation}. Code execution and pass rate are shown as percentages.}

\label{tab:gpt4o_agentic_performance}
\resizebox{1\columnwidth}{!}{%
\Large
\begin{tabular}{
    >{\raggedright\arraybackslash}p{4.4cm}
    | >{\raggedright\arraybackslash}p{3.3 cm}
    | c | c | c | c | c
}
\hline
\rule{0pt}{2.8ex}\textbf{Model Setup} & 
\rule{0pt}{2.8ex}\textbf{Strategy} & 
\makecell{\rule{0pt}{2.8ex}\textbf{Code Exec.} \\ \textbf{Success (\%)}} & 
\makecell{\rule{0pt}{2.8ex}\textbf{Answer} \\ \textbf{Match (\%)}} & 
\makecell{\rule{0pt}{2.8ex}\textbf{Clarity} \\ \textbf{Readability}} & 
\makecell{\rule{0pt}{2.8ex}\textbf{Chart} \\ \textbf{Correctness}} & 
\makecell{\rule{0pt}{2.8ex}\textbf{Final} \\ \textbf{Pass Rate (\%)}} \\
\hline
\multicolumn{7}{l}{\textit{\textbf{(A) Baseline}}} \\
\rowcolor[HTML]{E5F1FB} GPT-4o & 0-shot & 87 & 42 & 3.45 & 3.15 & 26 \\
\rowcolor[HTML]{E5F1FB} Gemini 1.5 Flash & 0-shot & 83 & 34 & 3.30 & 2.90 & 17 \\
\rowcolor[HTML]{E5F1FB} GPT-4o & 3-shot & 88 & 42 & 3.45 & 3.15 & 26 \\
\rowcolor[HTML]{E5F1FB} Gemini 1.5 Flash & 3-shot & 81 & 29 & 3.36 & 3.38 & 20 \\
\rowcolor[HTML]{E5F1FB} GPT-4o & RAG + 3-shot & 88 & 38 & 3.65 & 3.75 & 31 \\
\rowcolor[HTML]{E5F1FB} Gemini 1.5 Flash & RAG + 3-shot & 80 & 31 & 3.30 & 3.45 & 22 \\
\hline
\multicolumn{7}{l}{\textit{\textbf{(B) Agentic Inference (LLM Feedback)}}} \\
\rowcolor[HTML]{F3E8FD} GPT-4o + Gemini 1.5 & Answer + Code & 91 & 49 & 3.85 & 3.87 & 36 \\
\rowcolor[HTML]{F3E8FD} \textbf{GPT-4o + GPT-4o} & Answer + Code & \textbf{94} & \textbf{53} & \textbf{3.99} & \textbf{4.02} & \textbf{42} \\
\rowcolor[HTML]{F3E8FD} GPT-4o + GPT-4o & Answer + Code + Visual & 93 & 46 & 4.02 & 4.23 & 41 \\
\hline
\multicolumn{7}{l}{\textit{\textbf{(C) LLM Feedback Ablation}}} \\
\rowcolor[HTML]{E4F9E1} GPT-4o + GPT-4o & Answer Only & 86 & 47 & 3.51 & 3.20 & 28 \\
\rowcolor[HTML]{E4F9E1} GPT-4o + Matplotlib & Code Exec Only & 94 & 37 & 3.96 & 4.02 & 34 \\
\rowcolor[HTML]{E4F9E1} GPT-4o + GPT-4o & Code Only & 94 & 36 & 3.99 & 4.19 & 32 \\
\rowcolor[HTML]{E4F9E1} GPT-4o + GPT-4o & Visual Only & 94 & 38 & 4.03 & 4.24 & 33 \\
\hline
\end{tabular}
}
\end{table}

\vspace{-2mm}
\subsection{Human Evaluation}
\vspace{-2mm}
To validate the reliability of our automated evaluation, we manually annotated 236 samples stratified to reflect the full dataset’s distribution. Three representative models were assessed: GPT-4o (closed-source, high-performing), LLaMA-3.1-8B (open-source, lower-performing), and Qwen2.5-7B (open-source, best-performing), using the same rubric as the automated setup.
\begin{table}[t!]
\centering
\renewcommand{\arraystretch}{1.35} 
\vspace{-3mm}
\caption {Human Evaluation results on Text2Vis using direct inference for different models.}
\label{tab:model_performance_manual}

\Large 
\resizebox{1\columnwidth}{!}{
\begin{tabular}{>{\raggedright\arraybackslash}p{4cm} | c | c | c | c | c} 
\hline
\textbf{Model} & \makecell{\textbf{Code Exec.} \\ \textbf{Success (\%)}} & \makecell{\textbf{Answer} \\ \textbf{Match (\%)}} & \makecell{\textbf{Visual Clarity} \\ \textbf{Readability}} & \makecell{\textbf{Chart} \\ \textbf{Correctness}} & \makecell{\textbf{Final} \\ \textbf{Pass Rate (\%)}} \\
\hline
\rowcolor[HTML]{E5F1FB} \textbf{GPT-4o} & \textbf{87} & \textbf{39} & \textbf{3.32} & \textbf{3.30} & \textbf{30} \\
\rowcolor[HTML]{E5F1FB} Llama-3.1-8B & 72 & 28 & 1.79 & 1.67 & 9 \\
\rowcolor[HTML]{E5F1FB} Qwen2.5-7B & 80 & 31 & 3.03 & 2.94 & 17 \\
\hline
\end{tabular}%
}
\end{table}

Across all three models, automated and manual evaluations differed by under 15\% on all criteria:
\textit{Answer Match}, \textit{Readability}, and \textit{Chart Correctness}. We found strong correlation between human and model judgments for each metric, with high Pearson ($r$) and Spearman ($\rho$) coefficients (see Table~\ref{tab:correlation_analysis}). To further assess agreement on the final pass rate, we computed Cohen’s Kappa, obtaining an average score of $0.78$—indicating substantial alignment.

\paragraph{Evaluation Consistency and Robustness} To assess the generalizability of our evaluation framework, we included Gemini 1.5 Pro as an alternative judge alongside GPT-4o. Across all models, Gemini-based pass rates closely align with GPT-4o (Table~\ref{tab:text2vis_dual_finalrate}), with strong correlation between the two judges (Table~\ref{tab:gpt4o_vs_gemini_corr}) and comparable agreement with human annotations (Table~\ref{tab:correlation_analysis_gemini}).
Although both LLMs demonstrate high alignment with human judgments, GPT-4o achieved slightly better correlation and is more widely used as a standard evaluator. Accordingly, we report GPT-4o-based scores as the primary metric, while including Gemini results for robustness. We also conducted a repeatability study on the same stratified 236-sample subset  using GPT-4o, evaluating each query five times with identical prompts. Results showed minimal variation, with over 97.5\% of samples yielding consistent final pass rates. Further details (App. ~\ref{app:eval-robustness}).

\vspace{-1mm}

\vspace{-1 mm}

\subsection{Ablation Studies}
\vspace{-1mm}
We conducted ablation studies to assess feedback contributions in our agentic framework and analyze robustness across task types.
 \vspace{-2mm}
\paragraph{Feedback Modality Ablation.} As shown in Table~\ref{tab:gpt4o_agentic_performance}c, we compared full tri-modal feedback (answer, code, and visual) against reduced variants: answer-only, code-only, visual-only, and execution-based feedback using Matplotlib. Removing any modality significantly degraded performance, confirming that multi-modal integration is essential for effective agentic refinement. We also studied text-only and visual-only generation settings and found that jointly generating both answer and code does not degrade performance, see \ref{app:disjoint} and Tab \ref{tab:disjoint_ablation_results}.

\paragraph{Task-Type Robustness.}
We evaluated model performance across task categories (Table~\ref{tab:model_performance_breakdown}). Model performance declined for complex queries—especially those requiring web-based retrieval or multi-chart generation—across all metrics , reflecting LLM limitations in grounding external information and reasoning over interdependent outputs. GPT-4o and Gemini-1.5 Flash consistently outperformed open-source models on these tasks. Models also struggled with unanswerable queries, highlighting difficulties in recognizing when no valid visualization or response can be generated. Interestingly, they performed better on conversational queries, possibly due to the contextual grounding acquired during pre-training and instruction-tuning. Lastly, open-ended queries were handled more effectively than closed-ended ones, indicating a stronger ability to generate diverse and flexible responses.
\vspace{-2mm}
\section{Error Analysis}
\label{sec:error_analysis}
\label{sec:error_analysis}
\vspace{-1mm}

We conducted a qualitative error analysis by manually evaluating 200 randomly selected samples for each of the 11 models to identify key error patterns (see fig.\ref{fig:error_sum}). Our findings are as follows:

\textbf{Code Execution Errors} –
Common syntax and runtime failures such as unterminated strings, missing commas, shape mismatches (e.g., \textit{"shape mismatch: objects cannot be broadcast to a single shape"}), and attribute errors (e.g., \textit{"'PathPatch' object has no attribute 'get\_ydata'"}). Naming issues (e.g., \texttt{y} instead of \texttt{years}) and indentation errors also disrupted execution. See Figure~\ref{fig:error_types2}(c, d, f, g)


\textbf{Data Import and Retrieval Issues} – Frequent failures in defining in-context datasets (\textit{'df' is not defined}), handling non-standard date formats (\textit{time data 'Sept 2000' does not match \%b \%Y}), and executing web-based data retrieval. See Figure~\ref{fig:error_types2}h.

\textbf{Logical and Analytical Reasoning Errors} Mistakes in multi-step calculations, incorrect metric selection, and flawed logic led to misleading outputs. See Figure \ref{fig:error_types2}b.

\textbf{Visualization Clarity Issues} Issues like missing labels, inconsistent axis scaling, and poor color schemes impacted interpretability, even when technically correct. See Figure \ref{fig:error_types2}(e).

\textbf{Instruction-Following Failures} Several models ignored task constraints or attempted irrelevant actions. For example, CodeLlama-34B frequently used \texttt{pd.read\_csv('data.csv')} instead of processing the provided data. See Figure~\ref{fig:error_types2}a.

\textbf{Incomplete Code Generation}  Outputs from models like Mistral and LLaMA-3.1 were often missing dataset definitions, key methods or steps, resulting in unusable code. See Figure~\ref{fig:error_types2}g.

\vspace{-1mm}
\section{Conclusion}

We introduce \textbf{Text2Vis}, a benchmark for evaluating LLMs in text-to-visualization tasks,  featuring diverse datasets and over 20 chart types to support complex queries involving multi-step reasoning, retrieval, multi-chart generation, and conversations. Our evaluation of open- and closed-source models revealed critical limitations, with error analysis highlighting key areas for improvement. To address performance gaps, we propose a cross-modal agentic inference framework that enables LLMs to refine both textual answers and visualization code using structured feedback. This framework significantly enhances answer accuracy, chart correctness, and overall pass rates. In addition, we present a scalable and fully automated LLM-based evaluation pipeline that assesses correctness, code execution, and visual quality. 
We believe Text2Vis is a valuable resource for advancing LLM capabilities in visualization code generation, promoting better alignment with real-world tasks and enabling more accurate, high-quality visualizations.

\section*{Ethical Considerations}
Our work focuses on sharing benchmark data and evaluation results to promote transparency and reproducibility in text-to-visualization research. All datasets used in Text2Vis are publicly available. The authors manually verified all LLM-generated queries and visualizations to ensure data integrity and accuracy.

We maintained fairness in model comparisons by applying consistent evaluation criteria across both open-source and closed-source models. 

\section*{Limitations}
While Text2Vis provides a comprehensive benchmark for evaluating text-to-visualization generation models, it has some limitations. First, although our dataset incorporates diverse real-world and synthetic data, it may not fully capture the range of complexities found in specialized domains. Second, our evaluation relies on LLM-based automated assessment frameworks which, while efficient, may introduce biases in interpreting visualization quality or correctness. Although we tested with two different LLM-based judges and observed strong alignment with human evaluations as well as high correlation between the two judges, finer aspects of visualization aesthetics or interpretability may still be better assessed through manual review.

Additionally, our benchmark focuses on Python-based visualization using Matplotlib and Seaborn, which are widely adopted for their versatility, compatibility, and extensive support in data science workflows. While we do not provide native code for other frameworks like Vega-Lite or D3.js, the Python code can be readily converted to these libraries using existing LLMs or code converters. Future work may extend Text2Vis by enabling direct generation of Vega-Lite, D3.js, or Plotly code, allowing broader support for interactive and web-based visualizations.

Furthermore, while we describe our method as an agentic inference framework, the current implementation follows a fixed actor–critic loop, without dynamic planning or adaptive tool selection. Future extensions could incorporate a planning component that enables the model to autonomously decide when to revise, retrieve external information, or switch visualization libraries—moving closer to a truly autonomous agent architecture.

Lastly, while our agentic learning framework demonstrated significant improvements, it introduces computational overhead, which may limit scalability for larger datasets or more resource-constrained environments. As an alternative, we showed how Matplotlib feedback can also provide similar improvements in performance.

\bibliography{text2vis}
\newpage
\appendix
\section{Appendices}
\label{app:Appendice}

\subsection{Common Chart Types}
\label{app:chart_types}
Text2Vis includes a wide range of chart types that reflect the diversity of real-world data analysis tasks \ref{fig:chart_types}. Line charts are essential for visualizing trends over time, making them ideal for time-series forecasting and moving averages. Bar charts are widely used for comparing categorical variables, especially in demographic and economic data. Multichart visualizations simulate dashboard-style insights, often used in business analytics. Pie charts and donut charts are useful for showing part-to-whole relationships. Scatter plots and boxplots support statistical analysis such as correlation and distribution. Waterfall charts are often applied in financial reporting to track cumulative effects. By including this variety, Text2Vis ensures robust evaluation of models across diverse analytical scenarios.

\vspace{-1ex} %
\begin{figure}[t]
     \vspace{-3ex}
    \centering
\includegraphics[width=0.9\linewidth] {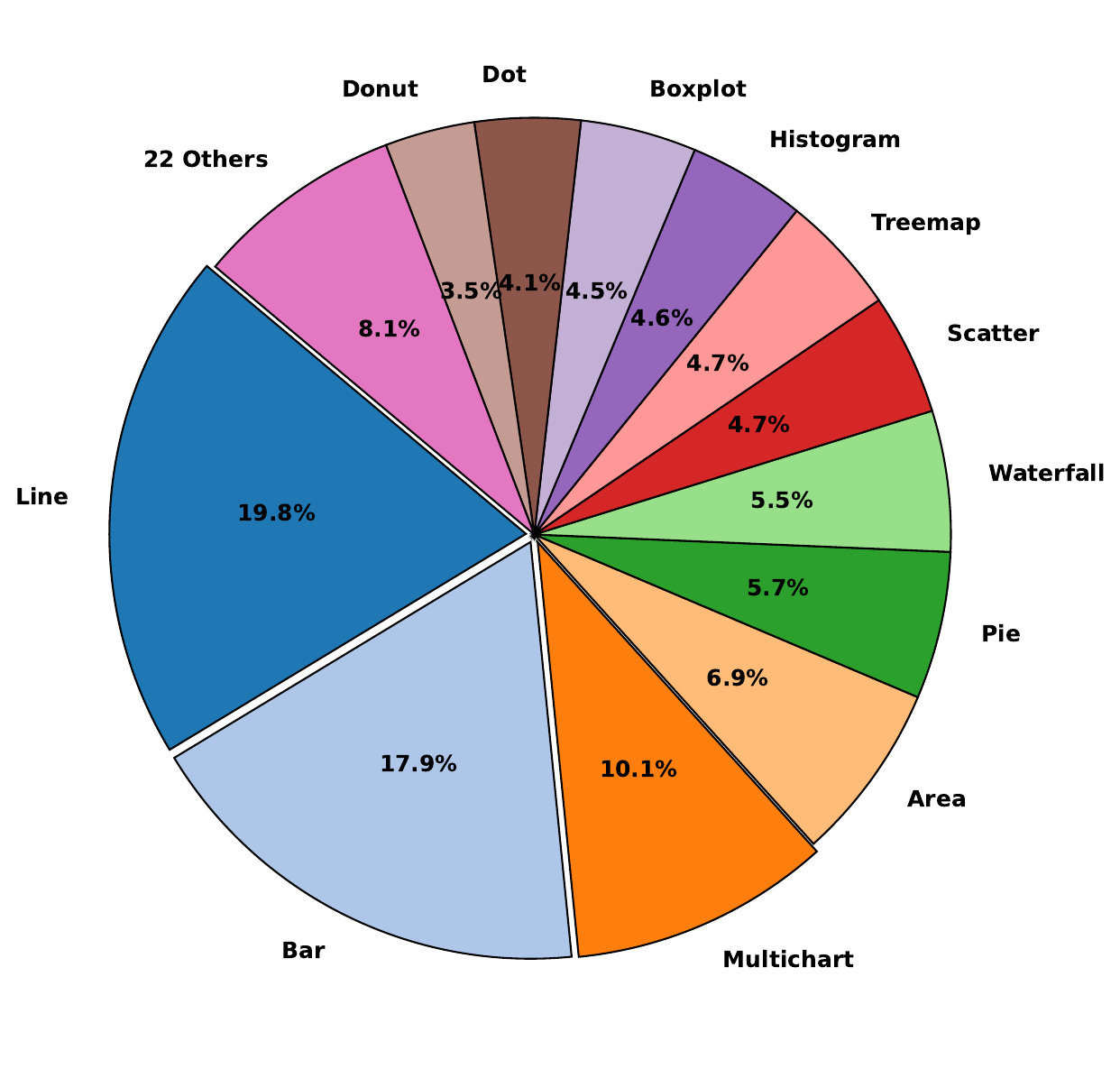}
\vspace{-4ex} %
    \caption{Common chart types in our Text2Vis. 
    }
    \label{fig:chart_types}
\end{figure}

\subsection{Data Science Question Taxonomy}
\label{app:framework}

Data science plays a crucial role in uncovering insights, identifying trends, making predictions, and driving informed decision-making. However, data-related questions vary in complexity and purpose. To better organize and analyze such questions, they can be categorized into four broad groups. These categories help structure the analytical approach and determine the appropriate methods for answering each type of question.

\subsubsection{Exploratory: 
Understanding Patterns and Structures}
\label{app:exploratory}
Some questions are aimed at understanding the overall structure of the data, identifying trends, or summarizing key characteristics. These questions do not necessarily seek to establish relationships between variables but rather focus on obtaining a broad overview of the dataset.

Exploratory questions can be categorized into the following subcategories: Insights, Trend Analysis, Statistical Summaries, Distribution Analysis, Categorical Data Analysis, Geospatial Analysis, Hierarchical Data Analysis, and Multi-Variable Analysis, each focusing on different aspects of understanding data patterns and structures.

\paragraph{Insights}
Insights-based questions focus on extracting key findings and meaningful observations from raw data. These questions often highlight notable patterns, distributions, or summary statistics.

\textbf{Example:} What are the top five best-selling products in the last six months?

\paragraph{Trend Analysis}
Trend analysis aims to identify changes in data over time, such as growth, decline, or seasonal fluctuations. These questions often involve historical patterns to detect trends.

\textbf{Example:} How have website visitor numbers changed over the past year?

\paragraph{Statistical Summaries}
Statistical summaries provide numerical insights into datasets, such as averages, variances, and standard deviations. These questions help quantify overall data characteristics.

\textbf{Example:} What is the median income of employees in each department?

\paragraph{Distribution Analysis}
Distribution analysis focuses on understanding how values in a dataset are spread across a range. It helps detect skewness, uniformity, or concentration in the data.

\textbf{Example:} What percentage of customers fall within different age groups?

\paragraph{Categorical Data Analysis}
These questions focus on analyzing groups of categorical variables to understand their distributions, relationships, or proportions.

\textbf{Example:} What percentage of total sales come from each product category?

\paragraph{Geospatial Analysis}
Geospatial analysis is concerned with visualizing and understanding spatial distributions across geographic regions.

\textbf{Example:} What is the distribution of customer locations by city?

\paragraph{Hierarchical Data Analysis}
Hierarchical analysis examines data structured in a nested or multi-level format, often represented through tree structures.

\textbf{Example:} How is the company's organizational hierarchy distributed across different departments?

\paragraph{Multi-Variable Analysis}
This analysis focuses on examining interactions between multiple variables simultaneously to identify complex relationships.

\textbf{Example:} How do age, income, and location influence customer purchasing behavior?

\subsubsection{Analytical: Explaining Relationships and Diagnosing Data}
\label{app:analytical}
Certain questions go beyond simple observation and focus on explaining why specific patterns or anomalies exist in the data. These questions investigate relationships between variables, detect irregularities, and provide insights into underlying factors.

Analytical questions can be categorized into the following subcategories: \textbf{Reasoning, Correlation Analysis, Outlier Detection, Deviation Analysis, and Comparison Analysis}, each focusing on uncovering relationships, detecting anomalies, and understanding variations in data.

\paragraph{Reasoning}
Reasoning-based questions focus on understanding causality, hypothesis testing, and logical deductions to explain why certain patterns or anomalies exist in the data.

\textbf{Example:} Why do customers in certain regions spend more on our products?

\paragraph{Correlation Analysis}
Correlation analysis examines the strength and direction of relationships between two or more variables, helping to understand dependencies in data.

\textbf{Example:} Is there a relationship between advertising budget and sales revenue?

\paragraph{Outlier Detection}
Outlier detection identifies unusual or extreme values in the dataset that may indicate errors, fraud, or unique trends.

\textbf{Example:} Are there any anomalies in the monthly transaction amounts that need investigation?

\paragraph{Deviation Analysis}
Deviation analysis measures how much data deviates from expected baselines, identifying significant variations or shifts in patterns.

\textbf{Example:} How much does employee performance vary from the expected target levels?

\paragraph{Comparison Analysis}
Comparison analysis focuses on evaluating differences and similarities between datasets, categories, or time periods.

\textbf{Example:} How do customer engagement metrics compare between last year and this year?

\subsubsection{Predictive: Forecasting Future Events}
\label{app:predictive}
Some questions are forward-looking, focusing on making informed predictions about future outcomes based on historical data. These questions rely on identifying past trends to estimate what is likely to happen next.

Predictive questions can be categorized into the following subcategories: \textbf{Predictive Analysis, Time-Series Analysis, Forecasting, and Anomaly Prediction}, each focusing on using past data to estimate future outcomes and detect potential irregularities.

\paragraph{Predictive Analysis}
Predictive analysis focuses on estimating future outcomes based on historical data patterns, often using statistical models or machine learning techniques.

\textbf{Example:} What is the likelihood that a customer will renew their subscription next year?

\paragraph{Time-Series Analysis}
Time-series analysis involves examining data that changes over time to identify trends, cycles, and seasonal effects.

\textbf{Example:} How do stock prices fluctuate over different time periods?

\paragraph{Forecasting}
Forecasting predicts future values based on past trends and patterns, commonly used in sales, finance, and demand planning.

\textbf{Example:} What will be the expected revenue for the next quarter?

\paragraph{Anomaly Prediction}
Anomaly prediction focuses on detecting rare but significant future events that deviate from expected patterns, such as fraud detection or equipment failures.

\textbf{Example:} Can we predict which transactions are likely to be fraudulent?

\subsubsection{Prescriptive: Recommending Data-Driven Actions}
\label{app:prescriptive}
Certain questions are designed to guide decision-making by providing actionable insights. Instead of just analyzing past data or predicting future trends, these questions focus on identifying the best possible course of action.

Prescriptive questions can be categorized into the following subcategories: \textbf{Decision Support, Classification \& Labeling, Clustering Analysis, and Causal Inference}, each focusing on recommending actions based on data insights and optimization techniques.

\paragraph{Decision Support}
Decision support focuses on recommending optimal strategies or actions based on data analysis. It helps businesses or individuals make informed choices by considering past trends and current conditions.

\textbf{Example:} What is the best pricing strategy to maximize profit while maintaining customer satisfaction?

\paragraph{Classification \& Labeling}
Classification and labeling involve assigning predefined categories or labels to new data points based on learned patterns from historical data.

\textbf{Example:} Should this email be categorized as spam or not?

\paragraph{Clustering Analysis}
Clustering analysis identifies groups of similar data points within a dataset, allowing segmentation and targeted decision-making.

\textbf{Example:} Can customers be segmented into different groups based on their purchasing behavior?

\paragraph{Causal Inference}
Causal inference seeks to determine cause-and-effect relationships between variables, helping understand the impact of changes or interventions.

\textbf{Example:} How does increasing the marketing budget affect customer retention rates?

\subsection{Query Types \& Illustrative Examples}
\label{app:query-types}

To ensure a comprehensive evaluation of text-to-visualization models, the Text2Vis dataset includes diverse query types that reflect real-world data analysis scenarios. These queries are designed to test various aspects of model reasoning, retrieval capabilities, response complexity, and visualization diversity. Specifically, we categorize our dataset along the following dimensions:
\begin{itemize}
    \item \textbf{Closed vs. Open-Ended Queries} – Distinguishes between questions expecting a specific answer as output (closed-ended) and the ones that are open-ended, allowing for multiple possible visualizations.
    \item \textbf{Single query vs. Conversational} – Differentiates between single query with multi-turn interactions where each query builds on prior responses and independent, standalone queries.
    \item \textbf{Data given vs. Web-data Retrieval} – Classifies queries based on whether they require retrieving external web data before generating visualizations.
    \item \textbf{Single vs. Multi-Chart} – Compares queries requiring a single visualization versus those needing multiple coordinated charts commonly found in dashboards and infographics.
    \item \textbf{Answerable vs. Unanswerable Queries} – Identifies whether a query has a definitive answer based on available data or if it requires additional assumptions, external knowledge, or subjective interpretation.
\end{itemize}


The following sections provide examples of  few of them.

\paragraph{Conversational Queries:} These queries simulate multi-turn interactions where each question builds on the previous answer, testing the model's ability to maintain context and continuity across queries.

\begin{itemize}
    \item \textbf{Q1:} Can you visualize the overall trend in unemployment in the USA from 2000 to 2020?  
    \textbf{A1 (Open-Ended):} The unemployment rate shows a significant spike during the 2008 financial crisis, peaking in 2009, followed by a steady decline until 2020.  
    \textbf{Code:} Line chart showing the unemployment trend from 2000 to 2020.

    \item \textbf{Q2:} What year had the highest unemployment rate?  
    \textbf{A2 (Short Answer):} 2009.  
    \textbf{Code:} Bar chart highlighting the year 2009 with the highest unemployment rate.

    \item \textbf{Q3:} Based on the provided unemployment trend graph, what key patterns and anomalies can you identify? Discuss any significant changes, potential causes, and long-term implications. 
    \textbf{A3 (Open-Ended):} The unemployment trend shows a sharp spike in 2009, likely reflecting the impact of the 2008 financial crisis. Post-2010, there is a gradual decline, suggesting economic recovery. However, smaller fluctuations in later years may indicate cyclical job market instabilities. A steep increase in recent years could be linked to external shocks such as a global pandemic or policy shifts.  
    \textbf{Code:} Line chart with an outlier marker on the year 2009.
\end{itemize}

\paragraph{Retrieval-Augmented Queries:} These queries require models to fetch additional data before visualization, testing their ability to integrate external data sources dynamically.

\begin{itemize}
    \item \textbf{Q1:} Retrieve the unemployment data for the USA from 2000 to 2020 and visualize the trend.  
    \textbf{A1 (Open-Ended):} The data shows a consistent trend with notable spikes during economic downturns, such as in 2009.  
    \textbf{Code:} Line chart showing the unemployment rate in the USA from 2000 to 2020 after retrieving relevant data.

    \item \textbf{Q2:} Based on the retrieved data, which year had the lowest unemployment rate?  
    \textbf{A2 (Short Answer):} 2019.  
    \textbf{Code:} Bar chart showing the year 2019 with the lowest unemployment rate.
\end{itemize}

\paragraph{Short Answer vs. Open-Ended Queries:} These queries distinguish between concise factual responses and detailed analytical insights.

\begin{itemize}
    \item \textbf{Short Answer Query:} What is the highest unemployment rate recorded in the USA between 2000 and 2020?  
    \textbf{A (Short Answer):} 9.6\% in 2009.  
    \textbf{Code:} Single bar chart highlighting 2009.

    \item \textbf{Open-Ended Query:} Analyze the unemployment trend in the USA from 2000 to 2020 and discuss any significant fluctuations.  
    \textbf{A (Open-Ended):} The data indicates a sharp rise in unemployment during the 2008 financial crisis, followed by a gradual recovery. The COVID-19 pandemic in 2020 caused another spike.  
    \textbf{Code:} Line chart with annotations on significant years (2009 and 2020).
\end{itemize}

\paragraph{Unanswerable Queries}

Unanswerable queries arise when the required data is not available in the dataset or the question cannot be logically answered based on the provided information. These queries generally fall into the following types:

\begin{itemize}
    \item \textbf{Missing Data Queries} – When the dataset does not contain the required information.  
    \textbf{Example:} Asking for unemployment data from 1995 when the dataset only covers 2000 onward.

    \item \textbf{Ambiguous Queries} – When the question lacks specificity and can have multiple interpretations.  
    \textbf{Example:} Asking for "employment trends" without specifying sector or region.

    \item \textbf{Contradictory Queries} – When the query asks for information that is logically impossible.  
    \textbf{Example:} Asking for the highest unemployment rate in 2025 when the dataset does not contain future data.

    \item \textbf{Hypothetical Queries} – When the question asks about alternative scenarios not represented in the data.  
    \textbf{Example:} Asking what the unemployment rate would have been if the 2008 financial crisis had not occurred.
\end{itemize}

\begin{table}[h]
\centering
\renewcommand{\arraystretch}{1.2} 
\setlength{\tabcolsep}{8pt} 
\caption{\small Distribution of various visualization tasks in the Text2Vis Dataset. Insights, Trend Analysis, Statistical Summaries, Distribution Analysis, Categorical Data Analysis, Hierarchical Data Analysis, and Multi-Variable Analysis fall under the Exploratory category. Reasoning, Correlation Analysis, Outlier Detection, Deviation Analysis, and Comparison Analysis fall under the Analytical category. Predictive Analysis, Time-Series Analysis, and Forecasting fall under the Predictive category. Decision Support falls under the Prescriptive category.}
\label{tab:ds_distribution}

\begin{tabular}{l c}
\toprule
\textbf{Question Type} & \textbf{Count} \\
\midrule
Comparison Analysis           & 467 \\
Deviation Analysis            & 362 \\
Trend Analysis                & 346 \\
Distribution Analysis         & 146 \\
Forecasting                   & 142 \\
Outlier Detection             & 140 \\
Statistical Summaries         & 138 \\
Correlation Analysis          & 75  \\
Reasoning                     & 52  \\
Predictive Analysis           & 31  \\
Hierarchical Data Analysis    & 22  \\
Time-Series Analysis          & 18  \\
Insights                      & 16  \\
Categorical Data Analysis     & 12  \\
Decision Support              & 7   \\

Others              & 11   \\
\bottomrule
\end{tabular}
\end{table}

\subsection{Zero/Few-Shot and Retrieval-Augmented Prompting}
\label{app:direct-inference-setups}

In our experiments, we employed three distinct setups for the direct inference approach to comprehensively evaluate model performance. Below, we provide detailed descriptions of each approach used:

\paragraph{(i) Zero-Shot Prompting:}
In the zero-shot prompting scenario, the model receives only the target query and the corresponding data table directly, without any preceding examples or additional context. The prompt explicitly instructs the model to generate the response in JSON format. This setup assesses the model's inherent ability to interpret and respond to queries based solely on its pre-trained knowledge and instruction-following capabilities.

\paragraph{(ii) 3-Shot Prompting:}
In the 3-shot prompting approach, the model is initially provided with three randomly selected examples, each consisting of a natural language query, the relevant data table, and the correct annotated answer. These three examples precede the actual target query and its associated data table. The purpose of this setup is to evaluate the model's capability to leverage few-shot learning, enabling it to infer and adapt patterns from a minimal number of illustrative examples before generating the response for the target query.

\paragraph{(iii) Retrieval-Augmented Generation (RAG) + 3-Shot Prompting:}
The RAG + 3-shot prompting approach further enhances the 3-shot method by dynamically retrieving relevant examples from the dataset based on semantic similarity to the target query. Specifically, we employed the Sentence-BERT model (\texttt{all-MiniLM-L6-v2}) to encode queries into embedding vectors. We then computed cosine similarity scores between the target query embedding and all available query embeddings within the dataset. The top three most semantically similar queries and their corresponding annotated answers and data tables were retrieved. These retrieved examples served as the few-shot context preceding the target query, providing contextually relevant information intended to improve the model's comprehension and response accuracy.

\begin{figure}[t]
    \centering
    \vspace{-6mm}
\includegraphics[width=0.9\linewidth] {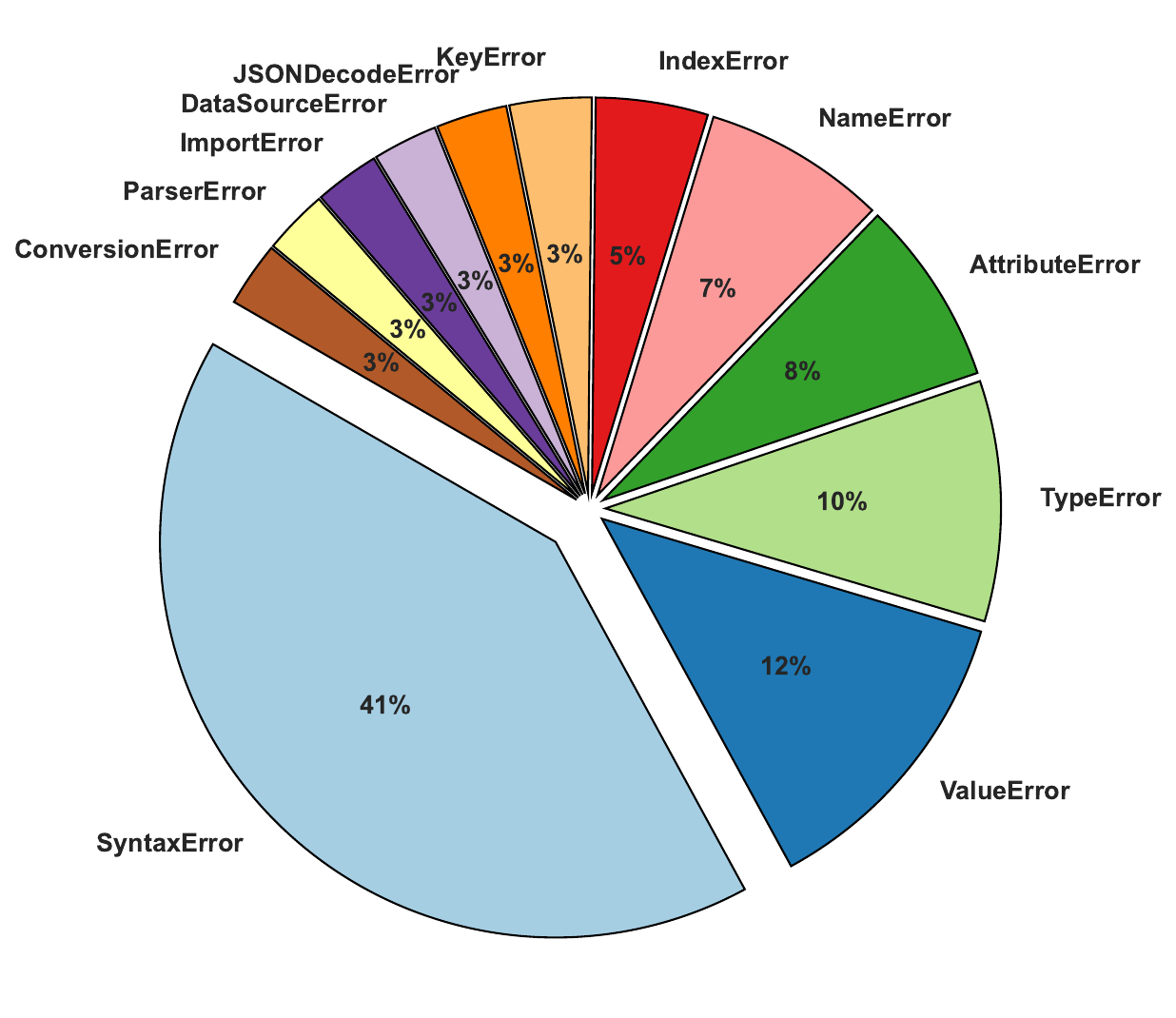}
\vspace{-6mm} %
    \caption{Error type distribution with square root transformation applied to prevent the SyntaxError category from dominating the chart. 
    }
    \label{fig:error_sum}
\end{figure}

\subsection{Common Data Visualization Error}
Figure~\ref{fig:error_sum} highlights examples of common visualization errors, including incorrect labeling, syntax errors, and data issues. Figure~\ref{fig:error_types2} also provides detailed examples of model failures. Finally, Figure~\ref{fig:error_types3} shows a word cloud of the most common words that appeared in our evaluation error messages.

\label{app:AMT}

\begin{figure*}[t]
    \centering
    \includegraphics[width=\textwidth]{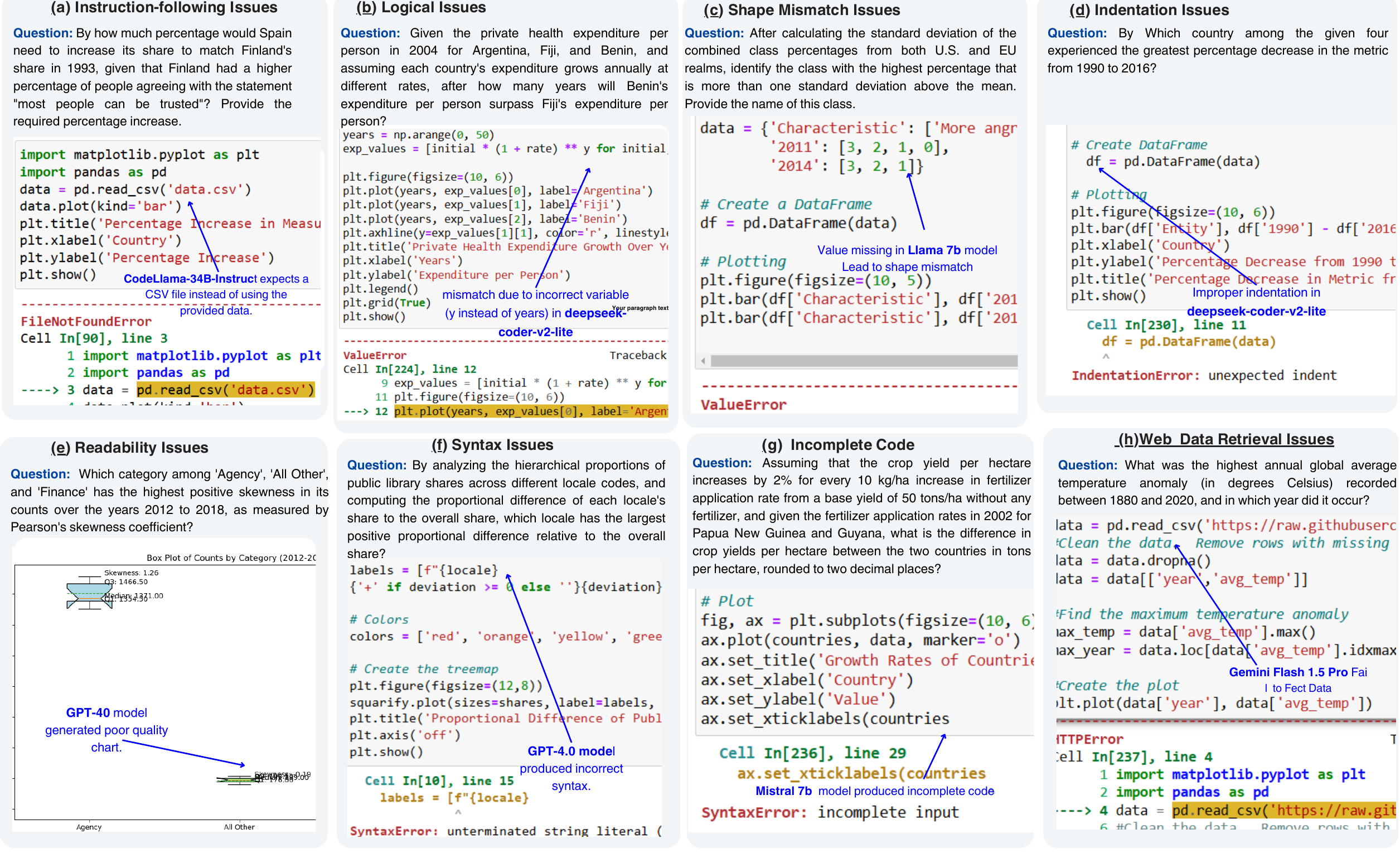}
    \caption{Common errors in Data Visualization generation.
    }
    \label{fig:error_types2}
\end{figure*}

\begin{figure*}[t]
    \centering
    \includegraphics[width=\textwidth]{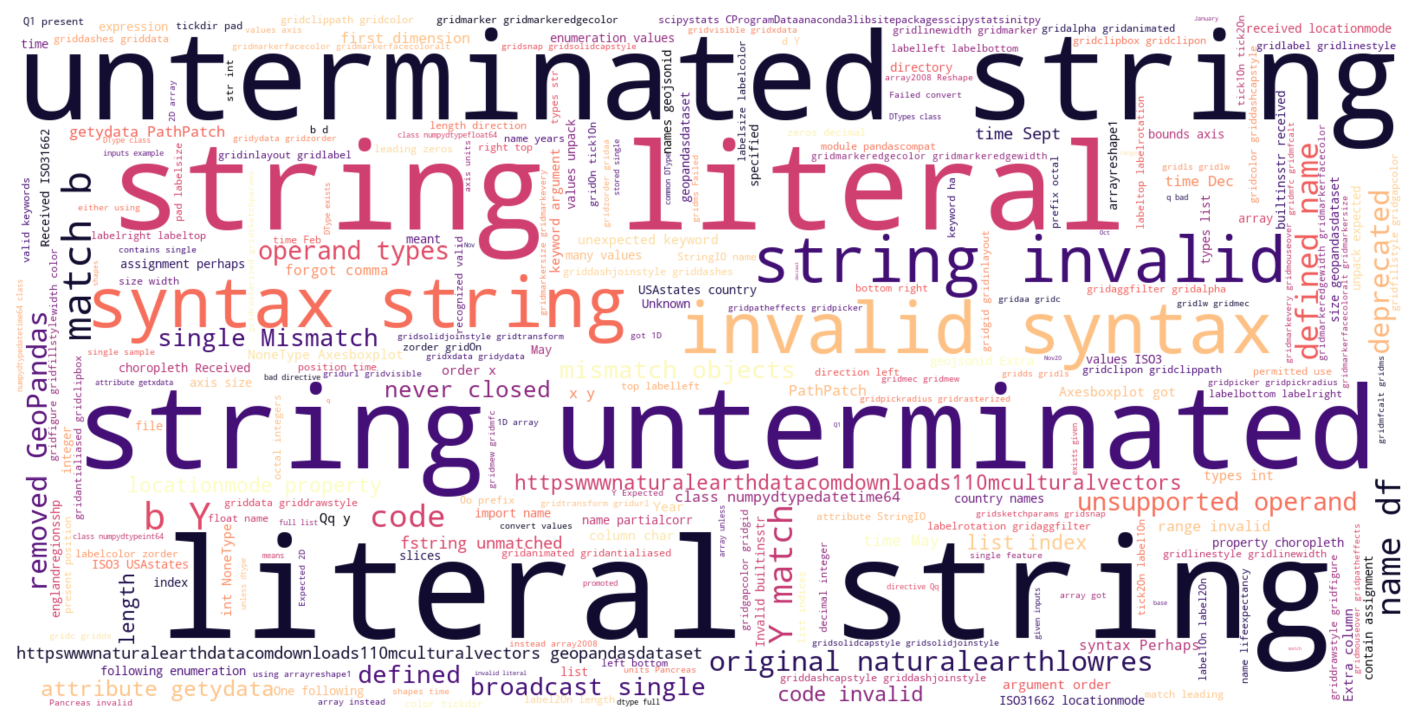}
    \caption{Most Frequent Words in Error Messages Across All Evaluated Models.
    }
    \label{fig:error_types3}
\end{figure*}

\definecolor{human_baseline}{RGB}{255, 245, 170}
\definecolor{open_models_below_4B}{RGB}{185, 235, 255}
\definecolor{open_models_7B_12B}{RGB}{255, 219, 187}
\definecolor{closed_models}{RGB}{240, 240, 240}
\definecolor{chart_specific_models}{RGB}{217, 240, 211}

\begin{table}[t!]
\vspace{-1mm}
\centering
\setlength{\tabcolsep}{1.7pt}
\renewcommand{\arraystretch}{1.25}
\caption{
Automatic evaluation results on Text2Vis using direct inference for different models. Higher values indicate better performance. Visual Clarity Readability and Chart Correctness are rated out of 5. Gemini 1.5 Pro as judge}
\label{tab:text2vis_direct_inference}
\resizebox{1\columnwidth}{!}{%
\begin{tabular}{>{\raggedright\arraybackslash}p{5cm}|c|c|c|c|c} 
\hline
\textbf{Model} & \makecell{\textbf{Code Exec.} \\ \textbf{Success (\%)}} & \makecell{\textbf{Answer} \\ \textbf{Match (\%)}} & \makecell{\textbf{Visual Clarity} \\ \textbf{Readability}} & \makecell{\textbf{Chart} \\ \textbf{Correctness}} & \makecell{\textbf{Final} \\ \textbf{Pass Rate (\%)}} \\
\hline
\rowcolor[HTML]{E5F1FB} \textbf{GPT-4o} & \textbf{87} & \textbf{40} & \textbf{3.38} & \textbf{3.10} & \textbf{24} \\
\rowcolor[HTML]{E5F1FB} Gemini-1.5-Flash & 83 & 34 & 3.35 & 2.69 & 15 \\ 
\rowcolor[HTML]{E5F1FB} CodeLlama-7b & 60 & 10 & 1.97 & 1.23 & 1 \\
\rowcolor[HTML]{E5F1FB} CodeLlama-13b & 52 & 14 & 1.90 & 0.90 & 3 \\
\rowcolor[HTML]{E5F1FB} CodeLlama-34b & 39 & 21 & 0.92 & 0.60 & 3 \\
\rowcolor[HTML]{E5F1FB} Llama-3.1-8B & 72 & 24 & 1.74 & 1.27 & 6 \\
\rowcolor[HTML]{E5F1FB} Mistral-7B & 39 & 22 & 1.45 & 1.10 & 4 \\
\rowcolor[HTML]{E5F1FB} Qwen2.5-7B & 80 & 28 & 2.86 & 2.40 & 12 \\
\rowcolor[HTML]{E5F1FB} Qwen2.5-Coder-7B & 31 & 22 & 1.30 & 1.08 & 3 \\
\rowcolor[HTML]{E5F1FB} DeepSeek-Coder-V2-Lite & 75 & 21 & 2.97 & 2.10 & 8 \\
\rowcolor[HTML]{E5F1FB} DeepSeek-R1-Distill-Llama-8B & 35 & 36 & 1.70 & 1.21 & 8 \\
\hline
\end{tabular}
}
\end{table}

\begin{table*}[t]
\centering
\renewcommand{\arraystretch}{1.35}
\setlength{\tabcolsep}{5pt}
\caption{
Pearson and Spearman correlations between the \textbf{GPT-4o judge} and the
\textbf{Gemini 1.5 Pro judge} across three evaluation dimensions:
Answer Match, Clarity \& Readability, and Chart Correctness.
}
\label{tab:gpt4o_vs_gemini_corr}

\resizebox{0.95\textwidth}{!}{%
\begin{tabular}{>{\raggedright\arraybackslash}p{4cm} |
                >{\centering\arraybackslash}p{1.8cm}
                >{\centering\arraybackslash}p{2.2cm}
                >{\centering\arraybackslash}p{2.2cm} |
                >{\centering\arraybackslash}p{1.8cm}
                >{\centering\arraybackslash}p{2.2cm}
                >{\centering\arraybackslash}p{2.2cm}}
\hline
\textbf{Model} & \multicolumn{3}{c|}{\textbf{Pearson}} & \multicolumn{3}{c}{\textbf{Spearman}} \\
\cline{2-7}
 & \textbf{Answer Match} & \textbf{Clarity \& Readability} & \textbf{Chart Correctness}
 & \textbf{Answer Match} & \textbf{Clarity \& Readability} & \textbf{Chart Correctness} \\
\hline
\rowcolor[HTML]{E5F1FB}\textbf{GPT-4o}                       & 0.93 & 0.83 & 0.90 & 0.93 & 0.71 & 0.87 \\
\rowcolor[HTML]{E5F1FB}\textbf{Gemini-1.5-Flash}             & 0.98 & 0.84 & 0.88 & 0.98 & 0.71 & 0.87 \\
\rowcolor[HTML]{E5F1FB}\textbf{CodeLlama-7B}                 & 0.97 & 0.77 & 0.63 & 0.97 & 0.74 & 0.67 \\
\rowcolor[HTML]{E5F1FB}\textbf{CodeLlama-13B}                & 0.97 & 0.95 & 0.77 & 0.97 & 0.95 & 0.83 \\
\rowcolor[HTML]{E5F1FB}\textbf{CodeLlama-34B}                & 0.97 & 0.97 & 0.84 & 0.97 & 0.99 & 0.92 \\
\rowcolor[HTML]{E5F1FB}\textbf{Llama-3.1-8B}                 & 0.96 & 0.96 & 0.83 & 0.96 & 0.97 & 0.91 \\
\rowcolor[HTML]{E5F1FB}\textbf{Mistral-7B}                   & 0.94 & 0.97 & 0.86 & 0.94 & 0.98 & 0.92 \\
\rowcolor[HTML]{E5F1FB}\textbf{Qwen-2.5-7B}                  & 0.95 & 0.95 & 0.84 & 0.95 & 0.85 & 0.86 \\
\rowcolor[HTML]{E5F1FB}\textbf{Qwen-2.5-Coder-7B}            & 0.95 & 0.96 & 0.88 & 0.95 & 0.97 & 0.87 \\
\rowcolor[HTML]{E5F1FB}\textbf{DeepSeek-Coder-V2-Lite}       & 0.95 & 0.94 & 0.79 & 0.95 & 0.83 & 0.83 \\
\rowcolor[HTML]{E5F1FB}\textbf{DeepSeek-R1-Distill-Llama-8B} & 0.91 & 0.84 & 0.89 & 0.91 & 0.85 & 0.88 \\
\hline
\end{tabular}}
\end{table*}

\begin{table*}[t]
\centering
\renewcommand{\arraystretch}{1.35}
\setlength{\tabcolsep}{5pt}
\caption{
Pearson and Spearman correlations between human evaluation and GPT-4o-based evaluation across three core dimensions: Answer Match, Clarity and Readability, and Chart Correctness.}
\label{tab:correlation_analysis}

\resizebox{0.95\textwidth}{!}{%
\begin{tabular}{>{\raggedright\arraybackslash}p{4cm} | 
                >{\centering\arraybackslash}p{1.8cm} 
                >{\centering\arraybackslash}p{2.2cm} 
                >{\centering\arraybackslash}p{2.2cm} |
                >{\centering\arraybackslash}p{1.8cm} 
                >{\centering\arraybackslash}p{2.2cm} 
                >{\centering\arraybackslash}p{2.2cm}}
\hline
\textbf{Model} & \multicolumn{3}{c|}{\textbf{Pearson}} & \multicolumn{3}{c}{\textbf{Spearman}} \\
\cline{2-7}
 & \textbf{Answer Match} & \textbf{Clarity \& Readability} & \textbf{Chart Correctness} 
 & \textbf{Answer Match} & \textbf{Clarity \& Readability} & \textbf{Chart Correctness} \\
\hline
\rowcolor[HTML]{E5F1FB} \textbf{GPT-4o} & 0.85 & 0.84 & 0.81 & 0.85 & 0.71 & 0.78 \\
\rowcolor[HTML]{E5F1FB} \textbf{Llama-3.1-8B} & 0.89 & 0.91 & 0.87 & 0.89 & 0.91 & 0.89 \\
\rowcolor[HTML]{E5F1FB} \textbf{Qwen2.5-7B} & 0.87 & 0.89 & 0.92 & 0.87 & 0.87 & 0.90 \\
\hline
\end{tabular}
}
\end{table*}

\begin{table*}[t]
\centering
\renewcommand{\arraystretch}{1.35}
\setlength{\tabcolsep}{5pt}
\caption{
Pearson and Spearman correlations between human evaluation and Gemini 1.5 Pro-based evaluation across three core dimensions: Answer Match, Clarity and Readability, and Chart Correctness.}
\label{tab:correlation_analysis_gemini}

\resizebox{0.95\textwidth}{!}{%
\begin{tabular}{>{\raggedright\arraybackslash}p{4cm} | 
                >{\centering\arraybackslash}p{1.8cm} 
                >{\centering\arraybackslash}p{2.2cm} 
                >{\centering\arraybackslash}p{2.2cm} |
                >{\centering\arraybackslash}p{1.8cm} 
                >{\centering\arraybackslash}p{2.2cm} 
                >{\centering\arraybackslash}p{2.2cm}}
\hline
\textbf{Model} & \multicolumn{3}{c|}{\textbf{Pearson}} & \multicolumn{3}{c}{\textbf{Spearman}} \\
\cline{2-7}
 & \textbf{Answer Match} & \textbf{Clarity \& Readability} & \textbf{Chart Correctness} 
 & \textbf{Answer Match} & \textbf{Clarity \& Readability} & \textbf{Chart Correctness} \\
\hline
\rowcolor[HTML]{E5F1FB} \textbf{GPT-4o} & 0.92 & 0.86  & 0.68  & 0.92  & 0.72 & 0.66 \\
\rowcolor[HTML]{E5F1FB} \textbf{Llama-3.1-8B} &0.91  &0.89  &0.83  &0.91  &0.90  &0.91  \\
\rowcolor[HTML]{E5F1FB} \textbf{Qwen2.5-7B} & 0.90  & 0.84 & 0.93  & 0.90  & 0.85  & 0.85  \\
\hline
\end{tabular}
}
\end{table*}

\begin{table*}[t]
\centering
\renewcommand{\arraystretch}{1.35} 
\setlength{\tabcolsep}{3pt}
\caption{
Performance breakdown (Final Pass Rate in percentages) for text-to-visualization models across different evaluation categories.
}
\label{tab:model_performance_breakdown}

\resizebox{1\textwidth}{!}{
\begin{tabular}{>{\raggedright\arraybackslash}p{5cm} | c | c | c | c | c} 
\hline
\textbf{Model} & 
\makecell{\textbf{Closed /} \\ \textbf{Open-Ended}} & 
\makecell{\textbf{Single Query /} \\ \textbf{Conversational}} & 
\makecell{\textbf{Data Given /} \\ \textbf{Web-data Retrieval}} & 
\makecell{\textbf{Single /} \\ \textbf{Multi-Chart}} & 
\makecell{\textbf{Answerable /} \\ \textbf{Unanswerable}} \\
\hline
\rowcolor[HTML]{E5F1FB} \textbf{GPT-4o} & \textbf{24} / \textbf{26} & \textbf{20} / \textbf{50} & \textbf{26} / 8 & \textbf{26} / \textbf{26} & \textbf{29} / 3 \\
\rowcolor[HTML]{E5F1FB} Gemini 1.5 Flash & 17 / 19 & 13 / 33 & 18 / \textbf{17} & 17 / 17 & 19 / 6 \\
\rowcolor[HTML]{E5F1FB} CodeLlama-7b-hf & 2 / 1 & 2 / 1 & 0 / 2 & 2 / 3 & 2 / 0 \\
\rowcolor[HTML]{E5F1FB} CodeLlama-13b-hf & 5 / 0 & 3 / 8 & 4 / 0 & 4 / 4 & 5 / 0 \\
\rowcolor[HTML]{E5F1FB} CodeLlama-34b-hf & 5 / 2 & 2 / 13 & 4 / 0 & 5 / 1 & 5 / 0 \\
\rowcolor[HTML]{E5F1FB} Llama-3.1-8B & 7 / 4 & 5 / 14 & 7 / 0 & 7 / 5 & 7 / 0 \\
\rowcolor[HTML]{E5F1FB} Mistral-7B & 5 / 9 & 4 / 12 & 4 / 6 & 6 / 4 & 6 / 1 \\
\rowcolor[HTML]{E5F1FB} Qwen2.5-7B & 3 / 15 & 11 / 22 & 14 / 0 & 14 / 6 & 14 / \textbf{7} \\
\rowcolor[HTML]{E5F1FB} Qwen2.5-Coder-7B & 4 / 4 & 2 / 11 & 14 / 0 & 4 / 3 & 4 / 0 \\
\rowcolor[HTML]{E5F1FB} DeepSeek-Coder V2-Lite & 10 / 9 & 8 / 21 & 10 / 2 & 10 / 9 & 11 / 4 \\
\rowcolor[HTML]{E5F1FB} DeepSeek-R1-Distill-Llama-8B & 6 / 10 & 6 / 10 & 7 / 2 & 7 / 5 & 7 / 2 \\
\hline
\end{tabular}%
}
\end{table*}

\subsection{Experimental Setup \& Inference Protocol}
\label{app:exp-setup}

\subsubsection{Model Parameters}
We used the following fixed seed and controlled decoding parameters for output generation, while all other settings were retained as default values provided by each model's API.
\begin{itemize}
    \item \textbf{Seed:} 42
    \item \textbf{Temperature:} 0.1
    \item \textbf{Top-p:} 1.0
    \item \textbf{Max New Tokens (Model Response Generation):} 2048
\end{itemize} 

\subsubsection{Inference Protocols}
We use three setups:
\begin{itemize}
    \item \textbf{Zero-Shot Prompting:} Query + table only, no examples.
    \item \textbf{Few-Shot Prompting (3-shot):} Three examples precede the query.
    \item \textbf{RAG + 3-Shot Prompting:} Three most semantically similar examples retrieved via SBERT, then appended before the query.
\end{itemize}

\subsubsection{Prompt Construction Summary}
For consistency, all prompting strategies use the same system template structure and JSON formatting requirements. Full prompt templates are listed in Appendix~\ref{app:prompt-templates}.


\subsection{Extended Evaluation \& Robustness} 

\subsubsection{Human Evaluation Setup}
\label{app:human-eval}

The 236  samples were first annotated by a primary annotator—an expert in data science and visualization. The annotations were then independently verified by a second annotator, with any disagreements resolved through discussion. Each model output was rated for answer correctness, chart readability, and visual accuracy using the same scoring rubric as in automated evaluation.

To quantify agreement between human and automated scores, we computed Pearson ($r$) and Spearman ($\rho$) correlations across all three core dimensions, (see Table \ref{tab:correlation_analysis}). The average Pearson correlations were $0.87$ for \textit{Answer Match}, $0.88$ for \textit{Clarity \& Readability}, and $0.867$ for \textit{Chart Correctness}. The corresponding Spearman correlations were $0.87$, $0.83$, and $0.857$, respectively. Additionally, we computed Cohen’s Kappa for agreement on final pass rate, obtaining an average $\kappa = 0.78$—indicating substantial alignment.

These findings support the alignment of automated and human judgments, reinforcing the robustness of our LLM-based evaluation framework.

\subsubsection{ LLM Judge Agreement}
\label{app:eval-robustness}

To evaluate consistency across LLM judges, we compared GPT-4o and Gemini 1.5 Pro on all benchmarked models. As shown in Table~\ref{tab:gpt4o_vs_gemini_corr}, the two judges exhibit strong agreement across all evaluation dimensions. The average Pearson correlations were $0.953$ for \textit{Answer Match}, $0.907$ for \textit{Clarity \& Readability}, and $0.828$ for \textit{Chart Correctness}. Similarly, the average Spearman correlations were $0.953$, $0.868$, and $0.857$ for the same dimensions, respectively.  Gemini also showed similarly strong correlations with human scores (Table~\ref{tab:correlation_analysis_gemini}), with average Pearson correlations of $0.91$, $0.863$, and $0.813$, and Spearman correlations of $0.91$, $0.823$, and $0.807$ for \textit{Answer Match}, \textit{Clarity \& Readability}, and \textit{Chart Correctness}, respectively. These results confirm the consistency and robustness of our automated evaluation framework across state-of-the-art LLMs.

\subsubsection{Evaluation Repeatability}
To test evaluation stability, we re-evaluated the same stratified 236-sample subset five times using GPT-4o with fixed prompts and identical sampling settings. The results showed minimal variation across runs, with standard deviations of $\pm$0.0 for \textit{Answer Match}, $\pm$0.015 for \textit{Readability}, $\pm$0.023 for \textit{Chart Correctness}, and $\pm$0.003 for the \textit{Final Score}. Notably, 97.49\% of the samples maintained the same final pass rate across all runs, confirming the reliability and consistency of our evaluation framework.

\subsubsection{ Disjoint vs Joint Generation (Ablation Study)} 
\label{app:disjoint}
We also ablated the task formulation by comparing \textit{text-only}, \textit{visual-only}, and \textit{joint (both)} generation settings using GPT-4o. As shown in Table~\ref{tab:disjoint_ablation_results}, the joint setup achieved similar performance to the disjoint settings across answer accuracy and chart quality metrics. These results suggest that combining answer and visualization generation does not degrade performance, supporting the feasibility of our unified task formulation.

\begin{table}[H]
\centering
\renewcommand{\arraystretch}{1.35} 
\vspace{-3mm}
\caption {Ablation study comparing GPT-4o performance in text-only, visual-only, and joint (both) generation settings.}
\label{tab:disjoint_ablation_results}

\Large 
\resizebox{0.9\columnwidth}{!}{
\begin{tabular}{>{\raggedright\arraybackslash}p{4cm} | c | c | c | c} 
\hline
\textbf{Setup} & \makecell{\textbf{Code Exec.} \\ \textbf{Success (\%)}} & \makecell{\textbf{Answer} \\ \textbf{Match (\%)}} & \makecell{\textbf{Visual Clarity} \\ \textbf{Readability}} & \makecell{\textbf{Chart} \\ \textbf{Correctness}} \\
\hline
\rowcolor[HTML]{E5F1FB} Text only & – & 44 & – & – \\
\rowcolor[HTML]{E5F1FB} Visual only & 89 & – & 3.61 & 3.30 \\
\rowcolor[HTML]{E5F1FB} Both & \textbf{87} & \textbf{42} & \textbf{3.45} & \textbf{3.15} \\
\hline
\end{tabular}%
}
\end{table}

\subsubsection{Runtime \& Cost Analysis} 
\label{app:cost}
To demonstrate scalability, we split the full 1,985-sample set into ten parallel GPT-4o API calls, completing all evaluations in just 5 minutes at a total cost of approximately \$2.0—compared to an estimated 33 human-hours (assuming 1 minute per sample)—yielding a significant reduction in annotation time and cost.

\subsection{Prompt Templates}
\label{app:prompt-templates}

We present the key prompt templates used throughout the Text2Vis. While we developed multiple query generation prompts to support various chart types and analytical tasks, representative examples are provided here.

\subsubsection{Query Generation Prompts}
To ensure diversity and realism in question types, we employ multiple prompts for structured query generation. A few representative examples include:

\begin{itemize}
    \item \textbf{Conversational Prompt:} Used to generate interdependent multi-turn queries over a dataset. See Table~\ref{tab:prompt_templat_conv}.
    \item \textbf{Histogram Prompt:} Generates synthetic frequency-based data and a complex reasoning question best answered with a histogram. See Table~\ref{tab:prompt_template_synthetic}.
    \item \textbf{Scatter Plot Prompt:} Used when the dataset contains multiple numerical columns, enabling queries about relationships and outliers. See Table~\ref{tab:prompt_template_scatter}.
\end{itemize}

Additional prompt types used during dataset creation include those for pie charts, bar charts, multi-chart dashboard layouts, and unanswerable question construction.

\subsubsection{Inference Prompts}
These prompts guide how models generate answers and visualizations, ensuring consistency across inference settings:

\begin{itemize}
    \item \textbf{Model Response Prompt:} A unified system prompt applied across zero-shot, few-shot, and RAG settings. It defines task instructions, output format (JSON), and required libraries (Matplotlib/Seaborn). See Table~\ref{tab:prompt_templat_response}.
    \item \textbf{Agentic Refinement Prompt:} Used in the self-refinement loop to assess and correct model outputs based on answer and chart quality. See Table~\ref{tab:agentic_framework}.
\end{itemize}

\subsubsection{Evaluation Prompts}
We adopt LLM-based evaluation prompts to assess answer and chart correctness and chart quality. Theis evaluates answer match, chart readability, and chart correctness using detailed rubrics. See Table~\ref{tab:result_evaluation}.
\subsubsection{Complexity Prompt}
This prompt is used to classify each query as \texttt{Simple}, \texttt{Medium}, \texttt{Hard}, or \texttt{Extra Hard}. See Table~\ref{tab:prompt_template_complexity}.

\begin{table*}[htbp]
  \centering
  \scriptsize
  \caption{Prompt Templates for Conversational Query Generation.}
  \label{tab:prompt_templat_conv}
  \rowcolors{2}{gray!10}{white}
  \begin{tabularx}{\textwidth}{>{\bfseries}l X}
    \toprule
    Category & Prompt Template \\ 
    \midrule
    \makecell[tl]{Conversational\\ Query \\ Generation}  & 
    \begin{minipage}[t]{\linewidth}
You are given a dataset in JSON format from my Data Table. Using this dataset, generate a \textbf{complex, conversational data analysis task} consisting of \textbf{4 to 5 interrelated steps}. Each step should logically build on the previous one to ensure a natural flow of analysis.\\[1ex]

To ensure clarity, \textbf{two examples} are included to demonstrate the expected structure. Please review these before generating new tasks. Then, create similar tasks that are \textbf{diverse, contextually relevant, and dependent on the new Data Table provided}.\\[1ex]

\textbf{Each conversation step should include:}\\[1ex]

\begin{itemize}
    \item \textbf{Question}: A data-driven question requiring multi-step reasoning (e.g., trend analysis, variability comparison, peak detection, forecasting) that directly relates to the dataset.
    \item \textbf{Answer}: Precisely answers the question.
    \item \textbf{Python Code Using Matplotlib}: A self-contained code snippet that generates a relevant visualization, including clear annotations highlighting key insights.
    \item \textbf{Text Summary}: A concise explanation of the insights derived from the visualization.
    \item \textbf{Metadata}: Include fields such as \texttt{"ChartType"}, \texttt{"xlabel"}, and \texttt{"ylabel"} to specify the visualization type and axis labels.
\end{itemize}

\textbf{Example Input:}\\[1ex]

Data Table

\texttt{...} \\[1ex]

\textbf{Expected JSON Output Format:}\\[1ex]

\texttt{
\{
  "Question": "...",
  "Answer": "...",
  "Code": "...",
  "TextSummary": "...",
  "ChartType": "",
  "xlabel": "...",
  "ylabel": "..."
\}
}\\

Ensure that each step builds on the previous one, creating a logically structured multi-step data analysis task. Maintain clarity, conciseness, and accuracy in all responses. Additionally, ensure that the generated tasks are diverse and well-aligned with the specific structure and patterns observed in the examples, while adapting to the new dataset provided.\\[1ex]
    \end{minipage} \\
    \bottomrule
  \end{tabularx}
\end{table*}
\begin{table*}[htbp]
  \centering
  \scriptsize
  \caption{Prompt Template for Synthetic Data Table and Histogram Question Generation}
  \label{tab:prompt_template_synthetic}
  \rowcolors{2}{gray!10}{white}
  \begin{tabularx}{\textwidth}{>{\bfseries}l X}
    \toprule
    Category & Prompt Template \\
    \midrule
    \makecell[tl]{Synthetic Data Table\\ and Histogram\\ Question Generation} &
    \begin{minipage}[t]{\linewidth}
    I want you to perform the following tasks:\\[1ex]

    1. \textbf{Generate a Synthetic Data Table}:\\
       - Randomly choose \textbf{one} domain from:  \\
         \textbf{Healthcare, Technology, Finance, Marketing, Retail, Education, Sports, Energy, Logistics, or any other relevant field}.\\
       - Clearly specify the selected \textbf{domain} in the output.\\
       - Ensure \textbf{domain-specific numerical scales} (e.g., revenue in 100k+, sales in thousands, ratings 1.0-5.0, percentages 0-100\%).\\
       - The dataset should include \textbf{5-8 rows} and \textbf{1 numerical column} with \textbf{frequency-based data} to represent a distribution.\\
       - The structure must \textbf{naturally lead to a complex data science question requiring a Histogram} to analyze \textbf{distribution, frequency, or variability}.\\
       - Return the dataset as a \textbf{valid JSON object} with column names as keys and values as lists.\\[1ex]

    2. \textbf{Generate a Single, Very Complex Data Science Question}:\\
       - The question must require \textbf{multi-step reasoning} and \textbf{deep analysis} related to \textbf{data distributions, frequency analysis, or variability}.\\
       - The question should involve \textbf{detecting patterns, finding skewness, assessing data spread, or identifying peaks and outliers}.\\
       - Ensure the question requires \textbf{multi-step computations} such as \textbf{mean, median, standard deviation, percentiles, or comparisons of histogram bins} before deriving the final answer.\\
       - \textbf{Do not modify the original dataset in any way (e.g., do not add missing values or create new data points).}\\
       - The question should be best answered using a \textbf{Histogram}.\\[1ex]

    3. \textbf{Provide a Short Answer}:\\
       - The answer must be \textbf{exactly one word or one number}.\\[1ex]

    4. \textbf{Output Python Code Using Matplotlib}:\\
       - The code should create a \textbf{Histogram} that effectively visualizes the dataset and addresses the generated question.\\
       - \textbf{Data should be binned appropriately to represent the distribution}.\\
       - \textbf{Ensure clear labeling of axes and meaningful annotations} to highlight key insights.\\
       - \textbf{Use colors, bin adjustments, and density plots if necessary} to enhance clarity.\\
       - \textbf{Must include text annotations} to enhance the chart’s clarity and insight.\\
       - You may use \textbf{pandas} for data handling if necessary.\\[1ex]

    5. \textbf{Include a Text Summary}:\\
       - Provide a concise summary.\\
       - Highlight the \textbf{main insight} derived from the visualization.\\[1ex]

    6. \textbf{Provide Metadata}:\\
       - \textbf{Domain}: The selected domain.\\
       - \textbf{ChartType}: Must be \texttt{"Histogram"}.\\
       - \textbf{xlabel}: The numerical variable representing the bins.\\
       - \textbf{ylabel}: The frequency count of values.\\[1ex]

    \textbf{Output Requirements}:\\[1ex]
    - Return \textbf{all} the above information in a \textbf{valid JSON format} without any additional text or commentary outside the JSON object.\\
    - Follow this exact JSON structure:\\[1ex]

    \texttt{\{}\\
    \ \ \texttt{"Domain": "..."},\\
    \ \ \texttt{"GeneratedDataTable": \{ "Value": [...] \},}\\
    \ \ \texttt{"Question": "..."},\\
    \ \ \texttt{"Answer": "..."},\\
    \ \ \texttt{"Code": "..."},\\
    \ \ \texttt{"TextSummary": "..."},\\
    \ \ \texttt{"ChartType": "Histogram"},\\
    \ \ \texttt{"xlabel": "..."},\\
    \ \ \texttt{"ylabel": "..."}\\
    \texttt{\}}\\[1ex]
    \end{minipage} \\
    \bottomrule
  \end{tabularx}
\end{table*}

\begin{table*}[htbp]
  \centering
  \scriptsize
  \caption{Prompt Template for Generating a Scatter Plot Query}
  \label{tab:prompt_template_scatter}
  \rowcolors{2}{gray!10}{white}
  \begin{tabularx}{\textwidth}{>{\bfseries}l X}
    \toprule
    Category & Prompt Template \\
    \midrule
    \makecell[tl]{Scatter Plot}  &
    \begin{minipage}[t]{\linewidth}
You are given the following data table:

\texttt{{data\_text}}

Before proceeding, evaluate whether the dataset is suitable for generating a question that is best answered by a scatter plot visualization. A dataset is considered suitable for scatter plot analysis if it contains at least two numerical variables that can be meaningfully compared.

If the dataset is \textbf{NOT} suitable for scatter plot analysis, please output an empty JSON object with the key \texttt{"skip"} set to \texttt{true} and do not generate any further content.

If the dataset is suitable, then perform the following tasks:

\begin{enumerate}
    \item \textbf{Generate a Single, Very Complex Data Science Question}:
      \begin{itemize}
          \item The question must require multi-step reasoning and deep analysis.
          \item Design the question specifically for a scatter plot visualization. For example, it may ask to analyze the relationship, correlation, or pattern between two numeric variables, identify outliers, or compare distributions.
      \end{itemize}
    \item \textbf{Provide a Short Answer}:
      \begin{itemize}
          \item The answer must be precise.
      \end{itemize}
    \item \textbf{Output Python Code for a Scatter Plot Visualization}:
      \begin{itemize}
          \item Use matplotlib to generate a scatter plot.
          \item Ensure the code annotates key insights on the plot.
      \end{itemize}
    \item \textbf{Include a Text Summary}:
      \begin{itemize}
          \item Provide a concise explanation of the reasoning behind the answer, highlighting the main insight derived from the scatter plot.
      \end{itemize}
    \item \textbf{Provide Metadata}:
      \begin{itemize}
          \item \textbf{ChartType}: Set this to \texttt{"Scatter"}.
          \item \textbf{xlabel}: The variable used for the X-axis.
          \item \textbf{ylabel}: The variable used for the Y-axis (if not applicable, use \texttt{"N/A"}).
      \end{itemize}
\end{enumerate}

To ensure clarity, two examples with scatterplot are included to demonstrate the expected structure. Please review these before generating new query and responses. Then, create similar query that are diverse, contextually relevant, and dependent on the provided data table.

\textbf{Output Requirements}:

\begin{itemize}
    \item Return all the above information in a \textbf{valid JSON format} without any additional text or commentary.
    \item Follow this exact JSON structure:
\end{itemize}

\textbf{Example Input:}\\[1ex]

Data Table

\texttt{...} \\[1ex]

\textbf{Expected JSON Output Format:}\\[1ex]

\texttt{
\{
  "Question": "...",
  "Answer": "...",
  "Code": "...",
  "TextSummary": "...",
  "ChartType": "Scatter",
  "xlabel": "...",
  "ylabel": "..."
\}
}\\
    \end{minipage} \\
    \bottomrule
  \end{tabularx}
\end{table*}

\begin{table*}[htbp]
  \centering
  \scriptsize
  \caption{Prompt Template for Model Response Generation}
  \label{tab:prompt_templat_response}
  \rowcolors{2}{gray!10}{white}
  \begin{tabularx}{\textwidth}{>{\bfseries}l X}
    \toprule
    Category & Prompt Template \\
    \midrule
    \makecell[tl]{Response }  &
    \begin{minipage}[t]{\linewidth}
You are a data visualization expert. Given a structured \textbf{data table}, respond to the following user question \textbf{based on the data}.

\textbf{Input Data:}  
\begin{itemize}
    \item \textbf{Data Table:} \{row['Data Table']\}  
    \item \textbf{Question:} \{row['Question']\}  
\end{itemize}

\textbf{Task:}
\begin{enumerate}
    \item \textbf{Answer}: Provide a precise and concise response based on the data. If no clear answer is available, return "unanswerable".  
    \item \textbf{Visualization Code}: Generate Python Matplotlib code to create a meaningful visualization that accurately represents the data. Ensure annotations and highlights are included.    
\end{enumerate}

\textbf{Important Requirement:}  
\begin{itemize}
    \item The output must be in a \textbf{valid JSON format} without any extra text, markdown formatting, or explanations.  
    \item Ensure the JSON structure strictly follows the format below.  
\end{itemize}

\textbf{Expected JSON Output Format:}
\begin{center}
\{
    "Answer": "...",
    "Visualization Code": "..."
\}
\end{center}
    \end{minipage} \\

    \bottomrule
  \end{tabularx}
\end{table*}

\begin{table*}[t]

  \centering
  \scriptsize
  \caption{Prompt Template for Agentic Framework} 
  \label{tab:agentic_framework}
  \rowcolors{2}{gray!10}{white}
  \begin{tabularx}{\textwidth}{>{\bfseries}l X}
    \toprule
    Category & Prompt Template \\
    \midrule
    \makecell[tl]{Agentic Framework}  & 
    \begin{minipage}[t]{\linewidth}
You are an expert in model response validation and refinement. Given a structured \textbf{data table}, Ground truth answer, a user-generated question, and an initial model response, your task is to validate and refine the model output for accuracy, correctness, and completeness.

\textbf{Input Data:}  
\begin{itemize}
    \item \textbf{Data Table}: \texttt{\{row['Table Data']\}}
    \item \textbf{Question}: \texttt{\{row['Question']\}}
    \item \textbf{Initial GPT-4o Response}: \texttt{\{gpt response\}}
\end{itemize}

\textbf{Task:}
\begin{enumerate}
    \item \textbf{Answer Validation}: Verify correctness and identify errors if any.
    \item \textbf{Visualization Code Validation}: Check for syntax errors, readability issues, or execution problems.
    \item \textbf{Refinement Task}:
        \begin{itemize}
            \item Based on the feedback, refine the model response to correct errors.
            \item Ensure the response is precise, formatted correctly, and adheres to the required JSON format.
        \end{itemize}
\end{enumerate}

\textbf{Output Requirements:}  
\begin{itemize}
    \item Ensure the final output is in a \textbf{valid JSON format} without extra text or markdown formatting.
    \item The JSON structure must strictly follow the format below.
\end{itemize}

\textbf{Expected JSON Output Format:}
\begin{center}
\{
    "Answer": "...",
    "Visualization Code": "..."
\}
\end{center}
    \end{minipage} \\

    \bottomrule
  \end{tabularx}
\end{table*}

\begin{table*}[htbp]
  \centering
  \scriptsize
  \caption{Prompt Template for Evaluating Results Using the GPT-4.0 Model. }
  \label{tab:result_evaluation}
  \rowcolors{2}{gray!10}{white}
  \begin{tabularx}{\textwidth}{>{\bfseries}l X}
    \toprule
    Category & Prompt Template \\
    \midrule
    \makecell[tl]{Evaluation}  & 
    \begin{minipage}[t]{\linewidth}
You are an evaluation expert responsible for assessing the accuracy of generated answers and the quality of visualizations. Given a structured \textbf{data table}, a user-generated question, a model-generated response, and an image-based visualization, your task is to validate the correctness of the response and evaluate the visualization quality.

\textbf{Input Data:}  
\begin{itemize}
    \item \textbf{Data Table}: \{\texttt{row['Table Data']}\}
    \item \textbf{Question}: \{\texttt{row['Generated Question']}\}
    \item \textbf{Generated Answer}: \{\texttt{row['Generated Answer']}\}
    \item \textbf{Ground Truth Answer}: \{\texttt{row['Answer']}\}
    \item \textbf{Generated Image}: \{\texttt{row['Generated image']}\}

\end{itemize}

\textbf{Task:}
\begin{enumerate}
    \item \textbf{Answer Matching}: Compare the generated answer with the ground truth using following evaluation  criteria.
    \item \textbf{Visualization Evaluation}: Score the visualization based on following evaluation  criteria.
\end{enumerate}

\textbf{Evaluation Criteria:}
\begin{enumerate}
    \item \textbf{Answer Matching (Binary: 1 or 0)}
    \begin{itemize} \tiny
        \item Match if numbers are close (e.g., "48.77" vs "48.73") or equivalent percentage formats (e.g., "100" vs "100
        \item Match if the ground truth appears within the generated response (e.g., "100" in "The result is 100").
        \item For long ground truth answer, match is considered as long as the core summary remains the same, even if the wording differs.
        \item Allow minor spelling variations or abbreviations (e.g., "Albenia" vs "Albania", "USA" vs "United States").
        \item No match if the meaning changes significantly (e.g., "Fragile" vs "Extreme fragility").
    \end{itemize}

    \item \textbf{Readability and Quality Score (0-5)}
    \begin{itemize} \tiny
        \item \textbf{Labels and Titles}: Are they clear, concise, and correctly positioned?
        \item \textbf{Layout Spacing}: Is the layout well-organized with no clutter?
        \item \textbf{Color Accessibility}: Are colors distinct and accessible (colorblind-friendly)?
        \item \textbf{Axis Scaling}: Are axes correctly labeled and proportional?
        \item \textbf{Chart Type Suitability}: Is the visualization appropriate for the data type (e.g., line chart for trends)?
        \item \textbf{Font and Legends}: Are fonts readable, and legends properly aligned?
        \item\textbf{Annotation Readability}: Are annotations (e.g., data labels, callouts) clear, well-placed, and non-overlapping?
    \end{itemize}

    \item \textbf{Chart Correctness Score (0-5)}
    \begin{itemize} \tiny
        \item \textbf{Query Alignment}: Does the visualization correctly address the question?
        \item \textbf{Data Integrity}: Are all data points accurately plotted?
        \item \textbf{Insight Representation}: Does the chart effectively communicate its key insights based on its type?
        \item \textbf{Handling Missing Data}: Is missing data presented appropriately without misleading distortion?
        \item \textbf{Complexity Handling}: For multi-step queries, is the visualization logically structured?
    \end{itemize}
\end{enumerate}

\begin{itemize}
    \item \textbf{5.0} – Excellent: Clear, accurate, and no issues.
    \item \textbf{4.5} – Very Good: Minor issues but does not impact understanding.
    \item \textbf{4.0} – Good: Small flaws like minor misalignments.
    \item \textbf{3.5} – Decent: Some readability/accuracy issues but still interpretable.
    \item \textbf{3.0} – Average: Noticeable problems that affect clarity or correctness.
    \item \textbf{2.5} – Below Average: Several issues that may lead to misinterpretation.
    \item \textbf{2.0} – Poor: Significant issues making the chart unclear.
    \item \textbf{1.5} – Very Poor: Major readability or correctness flaws.
    \item \textbf{1.0} – Unusable: Completely unclear or misleading.
    \item \textbf{0.0} – Failed: The visualization is unreadable or irrelevant.
\end{itemize}


\textbf{Output Requirements:}  
\begin{itemize}
    \item Ensure the final output is in a  valid JSON format without additional text.
\end{itemize}

\textbf{Expected JSON Output Format:}
\begin{center}
\{
    "Answer Match": "...",
    "Readability and Quality Score": "...",
    "Chart Correctness Score": "..."
\}
\end{center}
    \end{minipage} \\

    \bottomrule
  \end{tabularx}
\end{table*}

\begin{table*}[htbp]
  \centering
  \scriptsize
  \caption{Prompt Template for Question Complexity Classification}
  \label{tab:prompt_template_complexity}
  \rowcolors{2}{gray!10}{white}
  \begin{tabularx}{\textwidth}{>{\bfseries}l X}
    \toprule
    Category & Prompt Template \\
    \midrule
    \makecell[tl]{Complexity \\ Classification}  &
    \begin{minipage}[t]{\linewidth}
Prompt:  
You are given a data science question based on a table. Your task is to classify the complexity of the question into one of the following four categories:

\begin{itemize}
    \item \textbf{Simple}: The answer can be directly retrieved by locating a single value or label, with no calculation required.
    \item \textbf{Medium}: The question requires one or two reasoning steps, such as comparing values, calculating a difference or percentage, or sorting a small set of entries.
    \item \textbf{Hard}: The question involves multiple steps of reasoning—combining comparisons, aggregations, or filtering across rows or categories. It may require intermediate calculations to arrive at the answer.
    \item \textbf{Extra Hard}: The question demands complex, multi-step reasoning such as identifying trends, interpreting grouped patterns, performing advanced aggregations. It may also involve retrieving external information from the web to fully answer the query.
\end{itemize}

Based on the criteria above, classify the question using one of the following labels: \texttt{Simple}, \texttt{Medium}, \texttt{Hard}, or \texttt{Extra Hard}.  

Provide \textbf{only the label} as your final output.
    \end{minipage} \\
    \bottomrule
  \end{tabularx}
\end{table*}

\end{document}